\documentclass{article}

\usepackage{arxiv}

\usepackage[utf8]{inputenc} 
\usepackage[T1]{fontenc}    
\usepackage{hyperref}       
\usepackage{url}            
\usepackage{booktabs}       
\usepackage{amsfonts}       
\usepackage{nicefrac}       
\usepackage{microtype}      
\usepackage{lipsum}
\usepackage{graphicx}
\usepackage{hyperref}
\usepackage{epstopdf, epsfig}
\usepackage{bm}
\usepackage{amsthm}
\usepackage{amsmath,amssymb}
\usepackage[utf8]{inputenc}
\usepackage[T1]{fontenc}
\usepackage{mathptmx}
\usepackage[numbers]{natbib}
\usepackage{color}
\usepackage{subfigure}
\usepackage[ruled,linesnumbered]{algorithm2e}
\SetKwComment{Comment}{$\triangleright$\ }{}

\title{GrADE: A graph based data-driven solver for time-dependent nonlinear partial differential equations}

\author{
  Yash Kumar \\
  Department of Mechanical Engineering\\
  Delhi Technological University\\
  \texttt{yashk8481@gmail.com} \\
   \And
 Souvik Chakraborty \\
  Department of Applied Mechanics\\
  School of Artificial Intelligence (ScAI) \\
  India Institute of Technology (IIT) Delhi \\
  \texttt{souvik@am.iitd.ac.in} \\
}

\begin{document}
\maketitle

\begin{abstract}
The physical world is governed by the laws of physics, often represented in form of nonlinear partial differential equations (PDEs). Unfortunately, solution of PDEs is non-trivial and often involves significant computational time. With recent developments in the field of artificial intelligence and machine learning, solution of PDEs using neural network has emerged as a domain with huge potential. However, most of the developments in this field are based on either fully connected neural networks (FNN) or convolutional neural networks (CNN). While FNN is computationally inefficient as the number of network parameters can be potentially huge, CNN necessitates regular grid and simpler domain. In this work, we propose a novel framework referred to as the Graph Attention Differential Equation (GrADE) for solving time dependent nonlinear PDEs. The proposed approach couples FNN, graph neural network, and recently developed Neural ODE framework. The primary idea is to use graph neural network for modeling the spatial domain, and Neural ODE for modeling the temporal domain. The attention mechanism identifies important inputs/features and assign more weightage to the same; this enhances the performance of the proposed framework. Neural ODE, on the other hand, results in constant memory cost and allows trading of numerical precision for speed. We also propose depth refinement as an effective technique for training the proposed architecture is lesser time with better accuracy. The effectiveness of the proposed framework is illustrated using 1D and 2D Burgers’ equation. Results obtained illustrate the capability of the proposed framework in modeling PDE and its scalability to larger domains without the need for retraining. 
\end{abstract}


\section{\label{sec:level1}Introduction}

Many complex phenomena of scientific importance can be compressed into a few partial differential equations (PDEs). Solving them is key to understand these phenomena. Popular methods for solving PDEs include Finite Element Method \cite{zienkiewicz1977finite}, Finite Volume Method \cite{moukalled2016finite}, Finite Difference Method \cite{liszka1980finite}, and Boundary Element Method \cite{aliabadi2002boundary}. However, these methods are often computationally expensive and can take hours, if not days, to solve complex nonlinear PDEs on irregular domains. Therefore, even today, development of efficient methods for solving PDEs is a relevant problem.

With recent developments in the field of artificial intelligence and machine learning, data driven solution of PDEs has emerged as a possible alternative to the classical numerical techniques. The primary idea of these methods is to learn the dynamical evolution by using machine learning algorithms. Popular machine learning algorithms used for learning system dynamics include reduced-order models \cite{CHAKRABORTY201870, GAO}, polynomial chaos expansion \cite{PolyChaos}, and Gaussian
processes \cite{BILIONIS2013212, ATKINSON2019166}.
Others have tried to find governing equation using symbolic regression from data \cite{Schmidt81, Bongard9943}. \citet{Brunton3932} and \citet{RAISSI} also attempted to obtain equations that best describes the observed data. \citet{patel} used neural networks over fourier transforms regressing nonlinear operator in PDE.



In past decade, research has been focused around using FNN, RNN for incrementing dynamics of system. Popular approaches includes models based on Long Short Term Memories and transformers \cite{g_transformers, lstm_gan, rnn}. \citet{GENEVA} used physics constrained auto-regressive model for surrogate modeling of dynamical systems. Although these models have been effective at modeling the systems dynamics, there is a lack of transparency as they act as black box models. Moreover these are discrete models and predicts sequence separated with only fixed time step. {On the other hand, models used in \cite{WU20, Brunton3932} works with equation, but  requires numerical time derivative of data; this naturally becomes a potential source of error}.

Another popular class of methods for modeling nonlinear dynamical systems is the physics-informed neural network (PINN). The idea was initially applied to simple fully connected \cite{712178} and later extended to deep neural nets \cite{RAISSI_p}. The basic idea here is to place a neural network prior on the state variable and then estimate the neural network parameters by using a physics-informed loss function. Continuous time and discrete time formulations of PINN were proposed. PINN has been successfully applied to solve a wide array of dynamical systems including, but not limited to, fluid flow \cite{wessels2020neural}, heat transfer  \cite{cai2021physics}, fracture mechanics \cite{goswami2020transfer}, reliability analysis \cite{chakraborty2020simulation} and bio-mechanics \cite{liu2020generic}. Several improvements to the originally proposed PINN can also be found in the literature. For example, \citet{zhu2019physics} developed convolutional PINN for time-independent systems. \citet{GENEVA} developed an auto-regressive convolutional PINN for dynamical systems. A Bayesian variant of the same was also proposed. In both the works, discrete time variants of PINN were used. The primary advantage of PINN resides in the fact that no training data is needed.
However, the physics-informed loss function involved in PINN is difficult to optimize. Also, PINN assumes that the governing PDEs are exact, which is often not true. A few research directed towards addressing this issue can be found in the literature \cite{chakraborty2021transfer,meng2020composite}.



Success of ResNet \cite{targ2016resnet} in computer vision has attracted attention of researcher as it resembles Euler's time integration scheme and thus, introduces a bias in network architecture. Recently developed Neural ODE \cite{chen2018neural} extend this idea to more advanced integration schemes. These networks parameterize a differential equation as 
\begin{equation}
\label{eq:node}
    u_t = \bm f(x, t, u(x, t); \theta)
\end{equation}
where, dynamic function $\bm f: \mathbb{R} \times \mathbb{R} \to \mathbb{R}^n$ and initial value $y_0 \in \mathbb{R}^n.$
A NODE having one hidden unit encounters a problem where activation trajectories do not cross with depth of network limiting expressiblity of network. This is overcome by adding auxiliary dimensions \cite{ANODE}. There are various methods used for training NODE depending upon requirements. Besides usual auto-differentiation, adjoint-based back-propagation is used due to  being memory efficient, but requires more time steps. Some challenges are overcome by using checkpoint method \cite{adaptive_checkpoint, hybrid-checkpoint}.

In this work, we propose a novel framework, referred to as Graph Attention Differential Equation (GrADE) for learning system dynamics from data. The primary motivation behind GrADE resides in the fact that real-life data are unstructured and resides on irregular domains. This prohibits direct application of convolutional neural network based approaches.
GrADE combines Graph Neural Network (GNN) with Neural ODE. With GNN, one can easily handle unstructured data on a irregular domain. Within GrADE, GNN is used to model the spatial domain and Neural ODE is used to model the temporal domain. Among different GNN avaialble in the literature, we propose to use the graph attention (GAT) \citep{gat} within the proposed GrADE. Within GAT, attention mechanism is used on embedding of nodes during aggregation. Additionally, GrADE allows a streamlined way of embedding the boundary conditions of required solutions in the architecture of the graph connections on boundary nodes. This ensures the network prediction always meet the required boundary conditions. Example for the same are discussed in the paper. 

The rest of the paper is organized as follows. In Section \ref{sec:ps}, the problem statement has been defined. Brief review of fully connected neural network (FNN), attention mechanism, and GAT are presented in Section \ref{sec:sec3}. We present the proposed approach in Section \ref{sec:pa}. Section \ref{sec:ne} presents two examples to illustrate the applicability of the proposed approach. Finally, Section \ref{sec:conclusions} presents the concluding remarks.

\section{Problem statement} \label{sec:ps}

In this work, we are interested in discovering PDE using data.
Without loss of generality, we consider a system governed by the following system of PDEs
\begin{equation}
\label{eq:pde_sys}
\begin{aligned}
\bm u(\bm x, t)_t = \bm f(\bm x, \bm u(\bm x, t)),\ \bm x \in \Omega,\ t \in [0, T]\\
\bm B(\bm u) = \bm b(\bm x, t),\ \bm x \in \Gamma
\end{aligned}
\end{equation}
where $\bm u(\bm x, t) \in \mathbb{R}^{ndim}$ are the state variables, $\bm u(\bm x, t)_t$ is temporal derivative and $\bm B$ is operator for enforcing boundary conditions. $\bm x \in \Omega$ represents the spatial coordinates and $t \in [0,T]$ represents time. Initial state $\bm u(\bm x, 0)$ can be any real valued random field.

We assume that we have noisy measurements of the state variables at fixed time intervals. Data is of form 
$\mathcal D = \left\{ \bm U(\bm x, t_i) \right\}_{i=1}^{N_t}$ where, $N_t$ is the number of time-steps at which data is available.
$\bm U(\bm x, t_i)$ is vector representing the measurements of state at predefined fixed grid nodes at time $t_i$. 
We are interested in developing a framework that is able to predict the future evolution of the state variables $\bm u (\bm x, t)$ and $\bm { u} (\bm x, t)_t$. In other words, we are interested in learning the operator $\bm f \left(\cdot \right)$ in Eq. \eqref{eq:pde_sys}.

\noindent\textbf{Remark 1:} Time evolution of the state variable $\bm u (\bm x, t)$ can be easily learned by using ResNet \cite{qin2019data} or other similar framework. However, predicting time-evolution of $\bm {\ u} (\bm x, t)_t$ is non-trivial as no measurements for the same is available. One can always opt for time-derivative. Finite difference type schemes results in erroneous results because of the noise in the data.

\section{Brief review of feed-forward and graph neural network}\label{sec:sec3}
In this section, we briefly review the fundamentals of Feed-forward Neural Network (FNN), attention mechanism and Graph Attention (GAT). These three form the backbone of the proposed approach.
\subsection{Feed-forward neural network}
One of the key component of the proposed GrADE is fully connected feed-forward neural network (FNN) also known as multilayer perceptron. 
FNNs are universal approximator \citep{pinkus_1999}
and are extremely accurate in performing a wide array of tasks such as statistical pattern recognition, regression and classification. Consider a $\mathcal N_N:\mathbb R^{N_{in}} \mapsto \mathbb R^{N_{out}}$ to be a operator of a FNN. Considering $\bm x_{in} \in \mathbb R^{N_{in}}$ to be the input, the output $\bm x_{out} \in \mathbb R^{N_{out}}$ can be represented as
\begin{equation}\label{eq:nn}
    \bm x_{out} = \mathcal N_N(\bm x_{in}; \bm \theta),
\end{equation}
where $\bm \theta$ represents the parameters of the neural network operator $\mathcal N_N$. In essence, the neural network operator $\mathcal N_N$ is composition function of the form
\begin{equation}
\label{eq:fnn}
    \mathcal{N}(\cdot; \bm \theta) = (\sigma_M \circ \textbf{W}_{M-1}) \circ \cdot \cdot \cdot \circ (\sigma_2 \circ \textbf{W} ),
\end{equation}
where $\mathbf W_j$ is the weight matrix connecting layer $j$ and $(j+1)$, $\circ$ is operator composition, and $\sigma _j:\mathbb R \mapsto \mathbb R$ is the activation function corresponding to the $j-$th layer. Note that the activation function is applied on one component at a time.
The choice activation plays an important role in neural network. Popular activation functions available in the literature includes sigmoid, tan-hyberbolic, and rectified linear unit. Details on the activation function used in this paper is provided later.

For using a FNN in practice, one needs to estimate the parameters of $\mathcal N_N(\cdot;\bm\theta)$. This is generally achieved by maximizing the likelihood of the data (or minimizing an error function). Considering $\mathcal D_d = \left\{\bm x_{in}^{(i)}, \bm x_{out}^{(i)} \right\}_{i=1}^{N_d}$ to be the training data available, we can estimate the parameters $\bm \theta$ by minimizing the $\mathcal L_2$ loss-function,
\begin{equation}
    \bm \theta^* = \arg \min_{\bm \theta} \sum_{j=1}^{N_d}{\left\| \bm x_{out}^{(j)} - \mathcal N_N(\bm x_{in}^{(j)};\bm \theta) \right\|^2},
\end{equation}
where $\left\| \cdot \right\|$ represents the $\mathcal L_2$ norm. Note that other loss-functions like $\mathcal L_1$ norm can also be used.

\noindent \textbf{Remark 2:} FNN, although universal approximators, can potentially be computationally expensive. This is because all neurons at layer $j$ are connected to all neurons at layer $j+1$. Therefore, the number of parameters in FNN is quite high.

 
\subsection{Attention}
Another core component of the proposed GrADE is the attention mechanism. In a conventional neural network, the hidden activation function $\sigma \left(\cdot\right)$ acts on a linear combination of the input activation. For instance, if $\bm h_i$ is the hidden state and $\bm w_i$ represents the weight, in a conventional neural network, we have
\begin{equation}\label{eq:activation}
    \bm h_{i+1} = \sigma\left( \bm w_i^T \bm h_i \right),
\end{equation}
where $\bm h_{i+1}$ is the output of the $i-$th layer. Note that the $\bm w _i$ in Eq. \eqref{eq:activation} is constant. In attention mechanism, we take a different path where the weight vectors are dependent on the inputs. This is mathematically represented as
\begin{equation}\label{eq:attention_unit}
    \bm h_{i+1} = \sigma (\bm g (\bm h_i; \bm \alpha)^T \bm h_i), 
\end{equation}
where $\bm g\left(\cdot ; \bm \theta \right)$ represents a learnable function parameterized by parameters $\bm \alpha$. With such a setup, we are forcing the neural network to ``pay attention'' to different type of inputs in an adaptive manner.

The basic idea of attention was first proposed in the context of recurrent neural network (RNN); however the concept is equally applicable to other types of neural networks as well. For instance, \cite{bahdanau2014neural,luong2015effective} proposed soft attention mechanism where the context vector in the decoder function of RNN is allowed to be function of input encoding vectors. On the other hand, \cite{zheng2017learning} used attention mechanism within the convolutional neural network framework. Recently, PINN based on attention mechanism has also been developed \cite{rodriguez2021physics}. Motivated from \cite{gat}, we utilize attention mechanism within the graph neural network in this paper.

Researchers over the past few years have proposed different attention mechanism. For example, the attention mechanism shown in Eq. \eqref{eq:attention_unit} is a type of multiplicative attention unit. The most popular attention mechanism are perhaps the `dot product attention' and the `multi-head attention'. Transformers \cite{vaswani2017attention}, for instance, utilizes multi-head attention. Similarly, `soft' and `hard' attention mechanism can also be found in the literature. In hard attention, each output only attends one input location. However, with such a setup, the loss-function of neural network becomes non-differentiable. In this work, we use a soft attention mechanism. For further detauils on different attention mechanism, interested readers may refer \cite{vaswani2017attention,pml1Book}.


\subsection{Graph neural network with attention}
Having discussed the attention mechanism, we proceed to the last component of proposed GrADE, namely Graph Attention (GAT) \cite{gat}. We first briefly discuss graph neural network followed by GAT.

We define a graph $\mathcal G = \{ \mathcal V, \mathcal E \}$ having vertices $\mathcal V = \left\{ v_1, v_2, \ldots, v_N \right\}, N=\left| \mathcal V \right|$ and edges $\mathcal E \subseteq \mathcal V \times \mathcal V$.
An edge in a graph connects two vertices and is denoted as $e_{i,j}:=\left( v_i, v_j \right) \in \mathcal V \times \mathcal V$ with $1 \le i,j \le N$ and $i\ne j$. At this stage, we note that most real-world data lies on irregular domains (e.g., social networks, point cloud, biological networks) and can be represented using graphs. With graph neural network, it is possible to directly operate on these graphs. There are two major class of methods used for working with graphs. First is spectral methods which is based on spectral graph theory. It relies on convolution theorem for defining convolution on graph. Where Fourier transform of function on graph is performed via projecting the function on Fourier functions which are nothing but matrix of Eigenvectors of graph Laplacian obtained via expensive Eigen-decomposition. Moreover their is no guarantee of learning spatially localized filter. \citet{henaff2015deep} used linear combination of smooth kernels to approximate spectral filter resulting in localized spacial filters and smaller number of parameters. \citet{Cheb} and \citet{cayley} used Chebyshev and Cayley's expansion for estimating spectral filter, bypassing the expensive Eigen-decomposition. 
Second is spatial methods which applies convolution directly on graph. \citet{vanillagcn} introduced the vanilla GCNs in which nodes shared same weights with all neighbours. This approach can handle different neighborhood sizes and is independent of graph size. Recently developed GraphSAGE \cite{graphsage} differentiate between weights of central node from neighbours while sampling from neighborhood, this feature improves performance of the model over various inductive benchmarks. Unlike classic convolution nets, this setting is isotropic in nature and do not distinguish between neighbours. Anisotropy can be achieved naturally if we have edge feature or using mechanism differentiating neighbours. MoNets \cite{Monets} leverages Bayesian Gaussian mixture model parameters to differentiate between neighbours based on information about degree of node. GAT \cite{gat} used attention mechanism on node embeddings during aggregation.
 
Similar to convolutional neural networks, graph neural networks are composed of stacked layers, each performing message-passing and propagation. Most basic type of graph neural network layer can be represented vectorially as 
\begin{equation}
\label{eq:vanilaGNN}
    \bm h_i^{l+1} = \sigma (\frac{1}{d_i} \sum_{j \in N_i} \bm A_{ij} \bm W^l \bm h^l_j ),
\end{equation}
where, $\bm h^{l+1}_i \in \mathbb R^{d}$ has a dimensions of $d \times 1$. Eq. \eqref{eq:vanilaGNN} represents the operation performed on each node $v \in \mathcal V$ while implementing a layer of graph neural network. Note that the summation in Eq. \eqref{eq:vanilaGNN} is carried out over the $N_i$ neighbors of the $i-$th node.

\citet{gat} used attention mechanism for message passing on graph-structured data. This approach attend to neighbors based upon attention weights and were able to achieve state-of-the-art results on Cora, Citeseer and Pubmed citation network datasets. In GAT, Eq. \eqref{eq:vanilaGNN} is modified as follows
\begin{equation}
\label{eq:gnn_attention}
    \bm h^{l+1}_i =  \overset{K}{\underset{k=1}{\text{||}}} (\sigma(\sum_{j \in N_i} e^{k, l}_{ij} \bm W^{k,l} \bm h^l_j),
\end{equation}
where ${\text{||}}$ represents concatenation, $\bm W^{k,l}$ is the weight matrix for input linear transformation, and  $e^{k, l}_{ij}$ are the normalized attention coefficient and computed as
\begin{equation}\label{eq:attn_coeff}
    e_{ij} = \frac{\exp\left( \sigma ' \left( \bm \alpha ^T \left[ \bm W \bm h^l_i \text{||} \bm W \bm h^l_j \right]  \right) \right)}{\sum_{k\in N_i} \exp\left( \sigma ' \left( \bm \alpha ^T \left[ \bm W \bm h^l_i \text{||} \bm W \bm h^l_k \right]  \right) \right)}.
\end{equation}
Note that $\sigma '$ in Eq. \eqref{eq:attn_coeff} represents the activation functions and LeakyReLU is a popular choice in this case. $\bm \alpha$ in Eq. \eqref{eq:attn_coeff} represents the parameters of $\bm f \left(\cdot\right)$ defined in Eq. \eqref{eq:attention_unit}.



\section{Proposed approach}\label{sec:pa}
In this section, we discuss the proposed framework referred to here as Graph Attention Differential Equation (GrADE) for learning dynamics of systems from data. Recall that given data $\mathcal D = \left\{\bm x, t_i, \bm U(\bm x,t_i) \right\}_{i=1}^{N_t}$ at fixed time interval, the objective here is to learn the operator $\bm f\left(\cdot\right)$ in Eq. \eqref{eq:pde_sys}; this will allow predicting the state variable $\bm u(\bm x,t)$ and its derivative $\bm u(\bm x,t)_t$ at future time-steps. We note that unlike other similar works existing in the literature \cite{chen2018neural,queiruga2020continuous}, the state variable $\bm u$ for our case is dependent on both spatial location and temporal location and hence, both spatial and temporal discretization will be required.

We proceed by placing a neural network prior to parameterize the differential equation in Eq. \eqref{eq:pde_sys},
\begin{equation}\label{eq:net_estimate}
   \bm u_t = \mathcal N_N(\bm x, t, \bm u(\bm x, t); \bm \theta)
\end{equation}
where, $\bm u_t$ is time derivative of state vector $\bm u$, $\bm \theta$ are parameters of network $\mathcal N$. 
We rewrite Eq. \eqref{eq:net_estimate} as
\begin{equation}\label{eq:time_int}
    \bm u(\bm x, t_{k+1}) = \bm u\left(\bm x, t_k\right) + \int_{t_k}^{t_{k+1}} \mathcal N_N (\bm x, t, \bm u(\bm x, t)) dt.
\end{equation}
Eq. \eqref{eq:time_int} can be solved using some time integration scheme; although, time integration scheme introduces discretization error into the solution. For example, if we use Euler scheme, the approximation error is of the order $\mathcal O \left( \Delta t \right)$, where $\Delta t$ is the time-step. In this work, we have used fourth order Runge-Kutta (RK4 - 3/8) scheme,
\begin{subequations}\label{eq:RK4}
\begin{equation}
    \bm y_1 = \mathcal N_N\left(\bm x, t_k, \bm u\left(\bm x, t_k \right) \right),
\end{equation}
\begin{equation}
    \bm y_2 = \mathcal N_N\left(\bm x, t_k + \frac{\Delta t}{3}, \bm u\left(\bm x, t_k \right) + \left( \frac{\bm y_1}{3} \right) \right),
\end{equation}
\begin{equation}
    \bm y_3 = \mathcal N_N\left(\bm x, t_k + \frac{2\Delta t}{3}, \bm u\left(\bm x, t_k \right) - \left( \frac{\bm y_1}{3}  - \bm y_2 \right) \right),
\end{equation}
\begin{equation}
    \bm y_4 = \mathcal N_N\left(\bm x, t_k + \Delta t, \bm u\left(\bm x, t_k \right) + \left(\bm y_1 - \bm y_2 +\bm y_3 \right)\right).
\end{equation}
\begin{equation}\label{eq:RK4_pred}
    \bm u\left( \bm x, t_{k+1} \right) = \bm u \left( \bm x, t_k  \right) + \frac{\Delta t}{8} \left( \bm y_1 + 3\bm y_2 + 3\bm y_3 + \bm y_4 \right)
\end{equation}
\end{subequations}
We assume the system of interest is autonomous, i.e., $\mathcal N_N$ does not explicitly depend on the temporal variable $t$. Therefore, network will only take previous state and spatial coordinate as input and need not vary with depth. With slight abuse to terminology, we here refer to steps involved in time-integration scheme as depth.

\noindent \textbf{Remark 3:} The idea of parameterizing the differential equation by a neural network is motivated from Neural ODE \cite{chen2018neural} and continuous-in-depth network \cite{queiruga2020continuous}. However, both Neural ODE and continuous-in-depth network deals with ordinary differential equation. In our case the governing equation is a PDE.

To address the challenge mentioned in remark 3, we propose to use graph neural network to parameterize the operator $\bm f\left(\cdot\right)$ on the spatial domain. Accordingly, the neural network operator, $\mathcal N_N \left(\cdot\right)$ in Eq. \eqref{eq:RK4} is to be replaced with $\mathcal{GN}\left( \cdot \right)$. The advantage of graph neural network resides in the fact that, unlike FNN, it only utilizes information from neighboring nodes; this makes the model computationally tractable and scalable. Additionally, graph neural network also generalizes well on unseen spatial domains. To be specific, we design a custom graph attention (GAT) network for approximating the operator in the spatial domain, Details on the custom GAT network proposed in this paper are discussed next. 

\subsection{GAT architecture}
\begin{figure}[htb!]
        \centering
        \subfigure[]{
        \includegraphics[width = 0.37\textwidth]{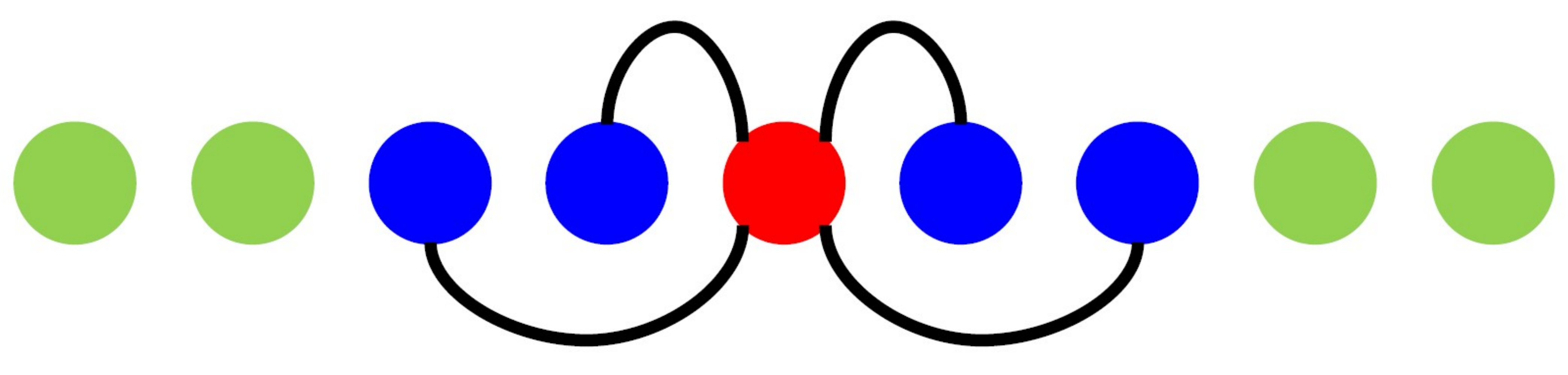}}\hspace{8mm}
        \subfigure[]{
        \includegraphics[width = 0.44\textwidth]{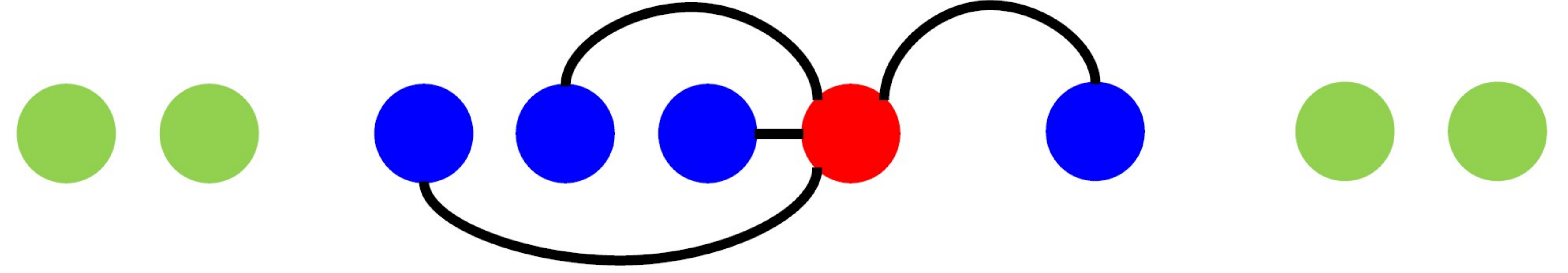}}
        \subfigure[]{
        \includegraphics[width = 0.25\textwidth]{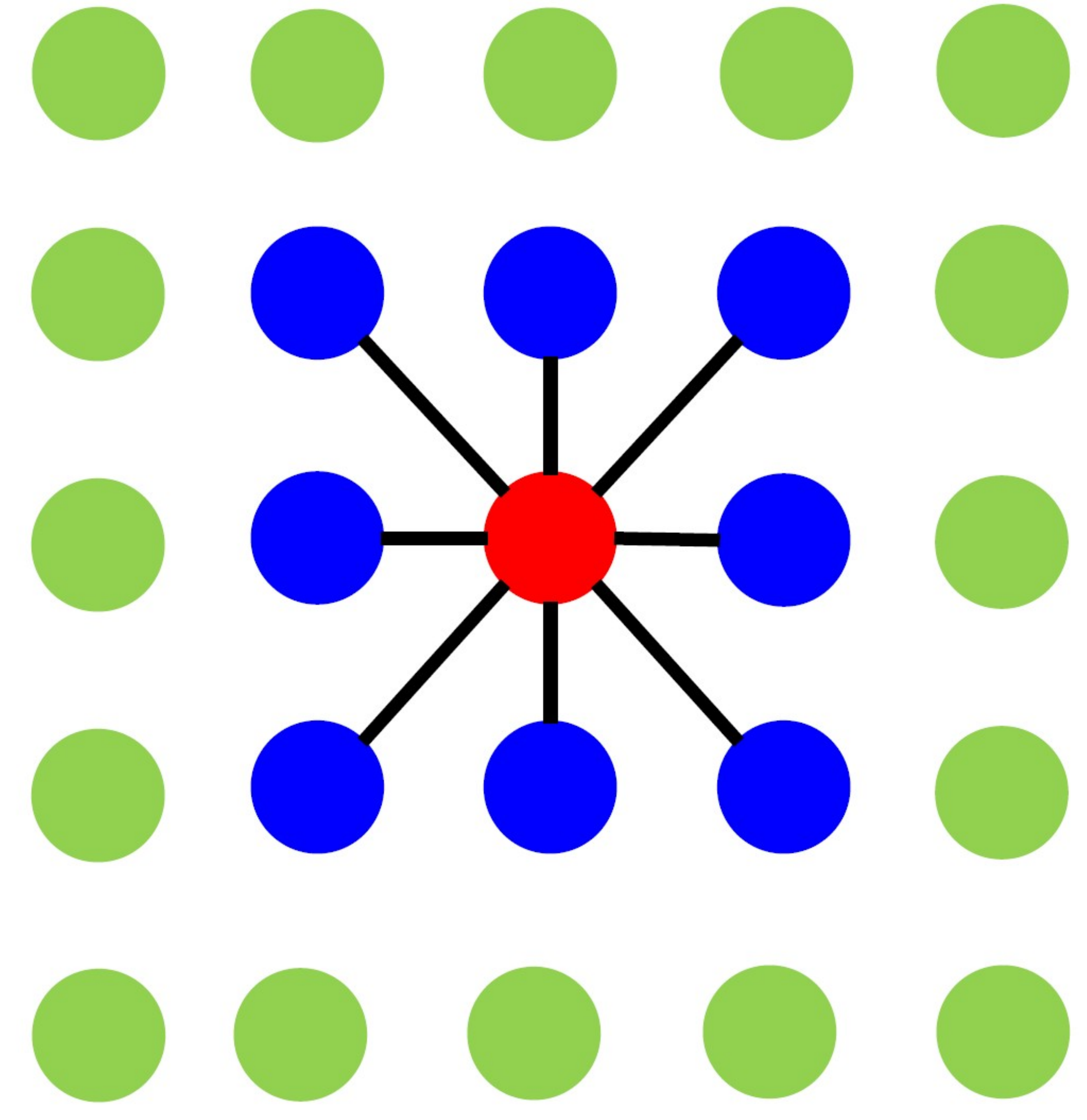}}\hspace{8mm}
        \subfigure[]{
        \includegraphics[width = 0.30\textwidth]{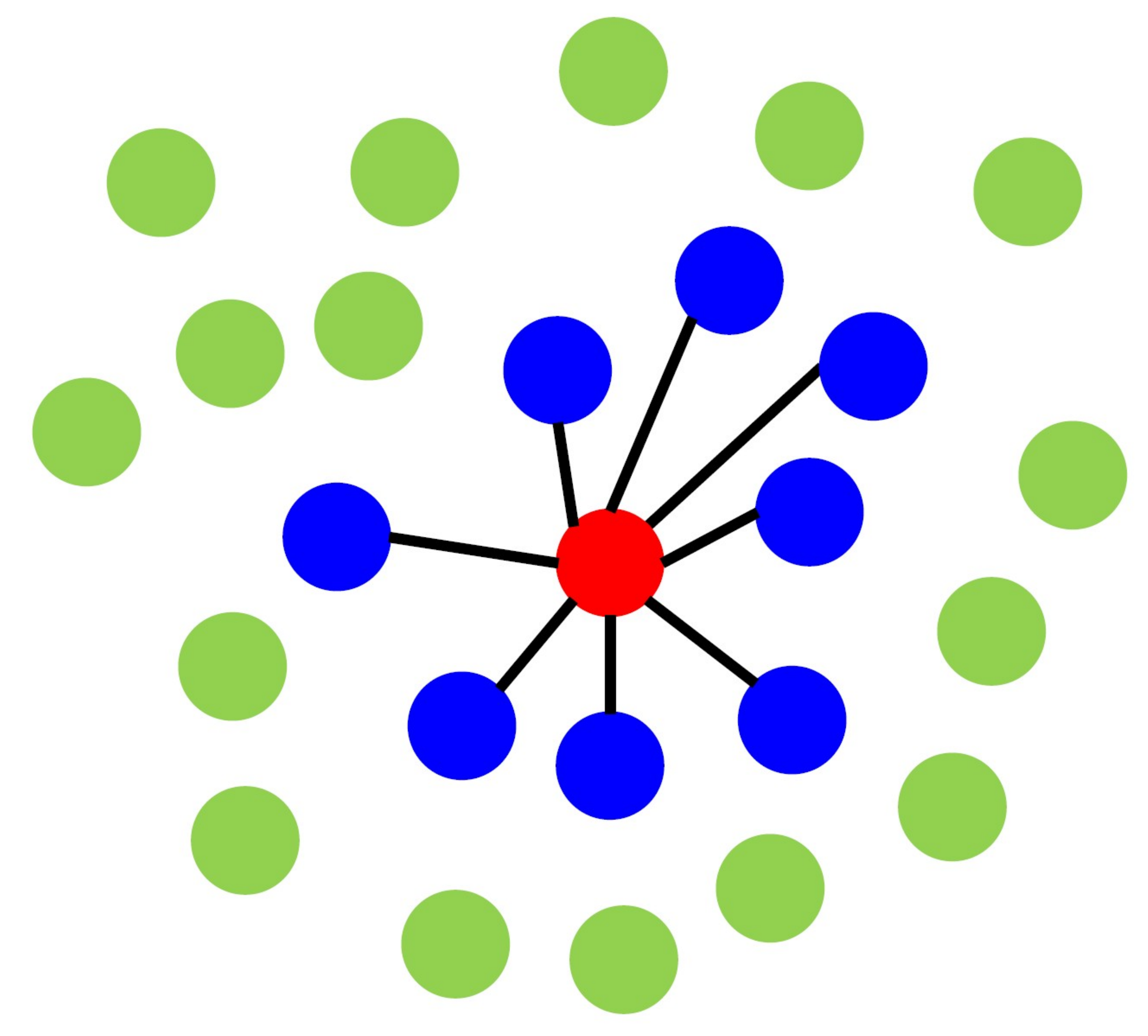}}
        \caption{Graphic showing connections with 4 and 8 nearest neighbor nodes in a 1D and 2D problem setting respectively. }
        \label{fig:graph_neighbor}
    \end{figure}

For approximating the operator $\bm f \left(\cdot \right)$ specified in Eq. \eqref{eq:pde_sys} in the spatial domain, we consider a network consisting of two graph network layers. For building the custom graph network, we consider the followings:
\begin{itemize}
    \item \textbf{Connection}: A node $v_i$ in the graph is connected to its neighboring nodes. For 1D problem, we consider 4 nearest nodes to be neighbors. Similarly for 2D problem, we consider 8 neighboring nodes to be neighbors. This is schematically shown in {Fig. \ref{fig:graph_neighbor}}. We use k-nearest neighbor algorithm \cite{peterson2009k} for determining neighbors of a node. Note that this will yield erroneous graph connection for the boundary nodes. In this work, the boundary nodes were modified manually.
    \item \textbf{Boundary conditions:} Boundary conditions (BC) are met by altering connection between graph nodes. e.g. for Dirichlet BC, we can remove the edges going towards the boundary nodes. This will prevent the value of boundary nodes from changing in next time step. We can specify new value for boundary at each time. Similarly for Neumann BC, we first remove edges going towards the boundary nodes and compute the boundary state variables based on the neighboring nodes by using the Taylors' series expansion. In this work, we consider examples with Periodic BC in which boundary nodes are connected both with local neighbours and nodes on opposite side of domain.
\end{itemize}

Once the graph network is designed using the method discussed above, we proceed with designing the architecture. This includes attention mechanism and different operations to be carried out within the network. Without loss of generality, let us consider the $i-$th node $v_i$ in a 2D graph. As per the rule discussed above, $v_i$ is connected to 8 neighbors. The edges connecting the nodes are denoted as $e_{i,k}:=\left(v_i, v_k\right), k=1,\ldots, 8$. For ease of understanding, we only focus on one edge $e_{i,j}$. We consider the spatial coordinates of the $i-$th and $j-$th nodes are  $\bm x_i = \left[x_i, y_i \right]$ and  $\bm x_j = \left[x_j, y_j \right]$, respectively. The graph architecture proposed in this work takes the relative difference between the spatial coordinates $\bm \delta \bm x_{i,j} = \left(\bm x_i - \bm x_j\right)$ and the  relative difference between the state variable $\bm \delta \bm u_{i,j} = \left(\bm u_i - \bm u_j \right)$ as inputs. We consider a case where network is composed of $2$ graph layers symbolised as $\mathcal{GN}^{(1)}$ and $\mathcal{GN}^{(2)}$.
First graph layer provides $\bm \delta \bm x_{i,j}$ as an input to a FNN $\mathcal N_{N_1}\left( \cdot; \bm \theta_N^{(1)} \right): \bm \delta \bm x_{i,j} \mapsto \bm \gamma_{i,j} $
\begin{equation}\label{eq:cust_GAT}
    {\bm \gamma_{i,j} = \mathcal N_{N_1}\left( \bm \delta \bm x_{i,j}; \bm \theta_{N}^{(1)} \right)},
\end{equation}
where $\bm \gamma_{i,j} = \left[ \gamma_{ux}^{(i,j)}, \gamma_{uy}^{(i,j)}, \gamma_{vx}^{(i,j)}, \gamma_{vy}^{(i,j)} \right] \in \mathbb R^4$ is the vector of attention weights for message $\mathcal M_{i, j}$ from neighbor $v_j$. $\bm \theta_{N}^{(1)}$ represents the parameters of the FNN.
%
%
The output of the FNN $\bm \gamma_{i,j}$ and the relative difference between the state variable $\bm \delta \bm u_{i,j}$ are then provided as an input to an operator $\bm H$ and the operator outputs the gradients of the state variable $\nabla \bm u_i$ at node $i$,
\begin{equation}\label{eq:H_operator}
    {\nabla \bm u_i = \sum_{j\in N_i} \bm H \left(\bm \gamma_{i,j}, \bm \delta \bm u_{i,j} \right) / N_i},
\end{equation}
where $N_i$ represents the neighbors of the $i-$th node. In essence, we design the operator $\bm H$ to first replicate $\delta \bm u_{i,j}$ as $\bm{\tilde u}_{i,j} = [\bm \delta \bm u_{i,j}, \bm \delta \bm u_{i,j}]$ and then carry out a Hadamard product with the neural network output,
\begin{equation}\label{eq:gradu}
\bm H \left(\bm \gamma_{i,j}, \delta \bm u_{i,j} \right) = \mathcal M_{i, j} = \bm {\tilde \gamma}_{i,j} \odot \delta \bm{\tilde u}_{i,j},
\end{equation}
where 
\begin{equation}
    \mathcal M_{i, j} = \left[ \begin{array}{cc}
    \left(u_x^{(i,j)}\right)_x & \left(u_y^{(i,j)}\right)_x  \\
     \left(u_x^{(i,j)}\right)_y & \left(u_y^{(i,j)}\right)_x
\end{array} \right], \; \bm {\tilde \gamma}_{i,j} = \left[ \begin{array}{cc}
    \gamma_{ux}^{(i,j)} & \gamma_{vx}^{(i,j)}  \\
     \gamma_{uy}^{(i,j)} & \gamma_{vy}^{(i,j)}
\end{array} \right], \; \text{and} \;  \delta \bm{\tilde u}_{i,j} = \left[ \begin{array}{cc}
   \delta u_x^{(i,j)} & \delta u_x^{(i,j)}  \\
     \delta u_y^{(i,j)} & \delta u_y^{(i,j)}
\end{array} \right].
\end{equation}
$\odot$ in Eq. \eqref{eq:gradu} denotes Hadamard product. Finally, we carry out a summation over the neighboring nodes to obtain the output of the first graph network layer. The basic premise here is that the first graph network layer, when trained, should output the gradients of the state vector $\nabla \bm u$. For sake of brevity, the overall operation carried out in the first layer of the graph network (at all nodes) is represented as
\begin{equation}\label{eq:GNN_layer1}
    \nabla \bm u = \mathcal{GN}^{(1)} \left( \bm x, \bm u; \bm \theta_N^{(1)} \right),
\end{equation}
with $\bm \theta_N^{(1)}$ being the parameters of the network.

The second graph network also functions in similar way as the first layer. Similar to the first graph network layer, we first provide the relative spatial coordinates as an input to a FNN $\mathcal N_{N_2} \left( \cdot; \bm \theta_{N}^{(2)} \right): \bm \delta \bm x_{i,j} \mapsto \bm \beta_{i,j} $
\begin{equation}\label{eq:NN2}
    \bm \beta _{i,j}= \mathcal N_{N_2}\left( \bm \delta \bm x_{i,j}; \bm \theta_{N}^{(2)} \right),
\end{equation}
where $\bm \beta_{i,j} = [ \beta_{uxx}^{(i,j)}, \beta_{uxy}^{(i,j)}, \beta_{uyx}^{(i,j)}, \beta_{uyy}^{(i,j)}, \beta_{vxx}^{(i,j)}, \beta_{vxy}^{(i,j)}, \beta_{vyx}^{(i,j)}, \beta_{vyy}^{(i,j)}] \in \mathbb R^8$ are the outputs from the FNN.
Again, this step represents the attention mechanism with $\bm \theta_{N}^{(2)}$ representing the neural network parameters. The output from $\mathcal N_{N_2} \left( \cdot; \bm \theta_{N}^{(2)} \right)$, $\bm \beta_{i,j}$ and the relative difference between the output from the first graph network layer are provided as inputs to the operator $\bm H $ and the operator outputs the second derivative of the state variables,
\begin{equation}\label{eq:cust_graph2}
    \mathbb H \left( \bm u_i \right) =  {\sum_{j\in N_i} \bm H \left( \bm { \beta}_{i,j}, {\delta{\nabla} {\bm u}_{i,j}}  \right)}.
\end{equation}
$ \mathbb H \left( \bm u_i \right) $ in Eq. \eqref{eq:cust_graph2} consist of the Hessian of the two state-variables, $u_x^{(i)}$ and $u_y^{(i)}$ at node $i$
\begin{equation}
    \mathbb H \left( \bm u_i \right) = \left[ \mathbb H_s \left( u_x^{(i)} \right), \mathbb H_s \left( u_y^{(i)} \right) \right],
\end{equation}
where
$\mathbb H_s \left( \cdot \right)$ represents the Hessian operator. Accordingly,
\begin{equation}\label{eq:hessian}
    \mathbb H_s \left( u_x^{(i)} \right) = \left[ \begin{array}{cc}
    \left( u_x^{(i)} \right)_{xx}     & \left( u_x^{(i)} \right)_{xy} \\
    \left( u_x^{(i)} \right)_{yx}     & \left( u_x^{(i)} \right)_{yy}
    \end{array} \right], \;\; \mathbb H_s \left( u_y^{(i)} \right) = \left[ \begin{array}{cc}
    \left( u_y^{(i)} \right)_{xx}     & \left( u_y^{(i)} \right)_{xy} \\
    \left( u_y^{(i)} \right)_{yx}     & \left( u_y^{(i)} \right)_{yy}
    \end{array} \right].
\end{equation}
and 
\begin{equation}\label{eq:hesian_comb}
     \mathbb H \left( \bm u_i \right) = \left[ \begin{array}{cccc}
       \left( u_x^{(i)} \right)_{xx}   & \left( u_x^{(i)} \right)_{xy} & \left( u_y^{(i)} \right)_{xx} & \left( u_y^{(i)} \right)_{xy} \\
       \left( u_x^{(i)} \right)_{yx}   &  \left( u_x^{(i)} \right)_{yy} & \left( u_y^{(i)} \right)_{yx}     & \left( u_y^{(i)} \right)_{yy}
     \end{array} \right]
\end{equation}
$\left(\cdot\right)_{kl}$ in Eqs. \eqref{eq:hessian} and \eqref{eq:hesian_comb} represents derivative with respect to variables $k$ and $l$. $\bm H \left( \tilde {\bm \beta}_{i,j}, \delta{\nabla} \tilde{\bm u }_{i,j} \right)$ in Eq. \eqref{eq:cust_graph2} is computed as
\begin{equation}\label{eq:H}
    \bm H \left( \tilde {\bm \beta}_{i,j}, \delta{\nabla} \tilde{\bm u}_{i,j} \right) = \tilde {\bm \beta}_{i,j} \odot \delta{\nabla} \tilde{\bm u}_{i,j},
\end{equation}
where $ \bm {\tilde \beta}_{i,j} \in \mathbb R^{2\times4}$ in Eq. \eqref{eq:H} is a matrix formulated by using the output of $\mathcal N_{N_2} \left( \cdot; \bm \theta_N^{(2)} \right)$
\begin{equation}\label{eq:beta_tilde}
    \bm {\tilde \beta}_{i,j} = \left[ \begin{array}{cccc}
        \beta_{uxx}^{(i,j)} & \beta_{uxy}^{(i,j)}  & \beta_{vxx}^{(i,j)} & \beta_{vxy}^{(i,j)}\\
         \beta_{uyx}^{(i,j)} & \beta_{uyy}^{(i,j)}  & \beta_{vyx}^{(i,j)} & \beta_{vyy}^{(i,j)}
    \end{array} \right].
\end{equation}
$ \delta_\nabla {\tilde{\bm u }}_{i,j}$ in Eq. \eqref{eq:H} is formulated by first creating a copy of each column of  $\delta_\nabla \bm u_{i,j}$, which in turn is computed by using the output of the first graph network layer
\begin{equation}\label{eq:u_tilde}
     \delta{\nabla}{ \tilde {\bm u}_{i,j} } = \left[ \begin{array}{cccc}
        \delta \left[\left( u_x^{(i,j)} \right)_x \right] & \delta \left[\left( u_x^{(i,j)} \right)_x\right] & \delta \left[\left( u_y^{(i,j)} \right)_x\right] &   \delta \left[\left( u_y^{(i,j)} \right)_x\right]  \\
        & & & \\
         \delta \left[\left( u_x^{(i,j)} \right)_y\right] & \delta \left[\left( u_x^{(i,j)} \right)_y\right] & \delta \left[\left( u_y^{(i,j)} \right)_y\right] &  \delta \left[\left( u_y^{(i,j)} \right)_y\right]
    \end{array} \right],
\end{equation}
where
\begin{equation}
    \delta\left[ \left(u_k^{(i,j)}\right)_l \right] = \left(u_k\right)_l^{(i)} - \left(u_k\right)_l^{(j)},\;\;k,l=x,y.
\end{equation}
$i$ and $j$ in superscript denotes the $i-$th and $j-$th nodes in the graph connected through edge $e_{i,j}$. For sake of brevity, we represent the overall operation carried out in second graph network layer as
\begin{equation}
    \mathbb H \left( \bm u \right) = \mathcal {GN}^{(2)} \left( \bm x, \nabla \bm u; \bm \theta_N^{(2)} \right).
\end{equation}

\begin{figure}[ht!]
    \centering
    \subfigure[GrADE]{\includegraphics[width=0.70\linewidth]{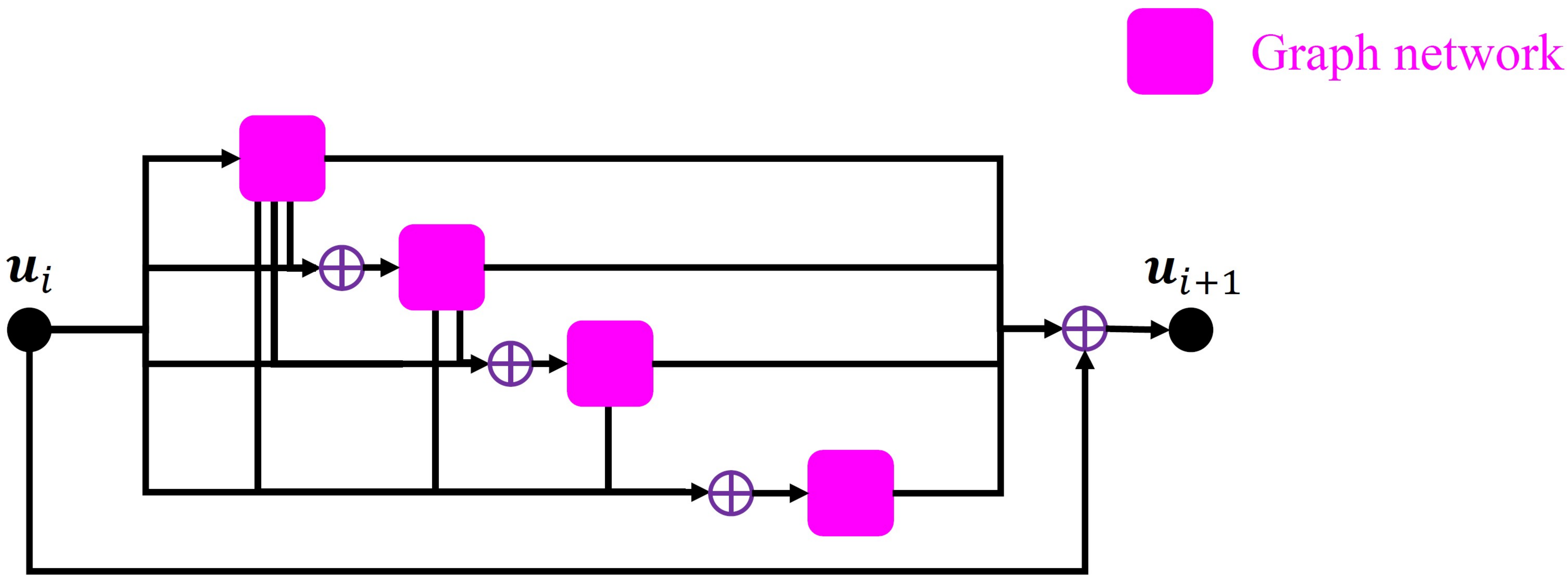}}
    \subfigure[Graph network (1D)]{\includegraphics[width=0.75\linewidth]{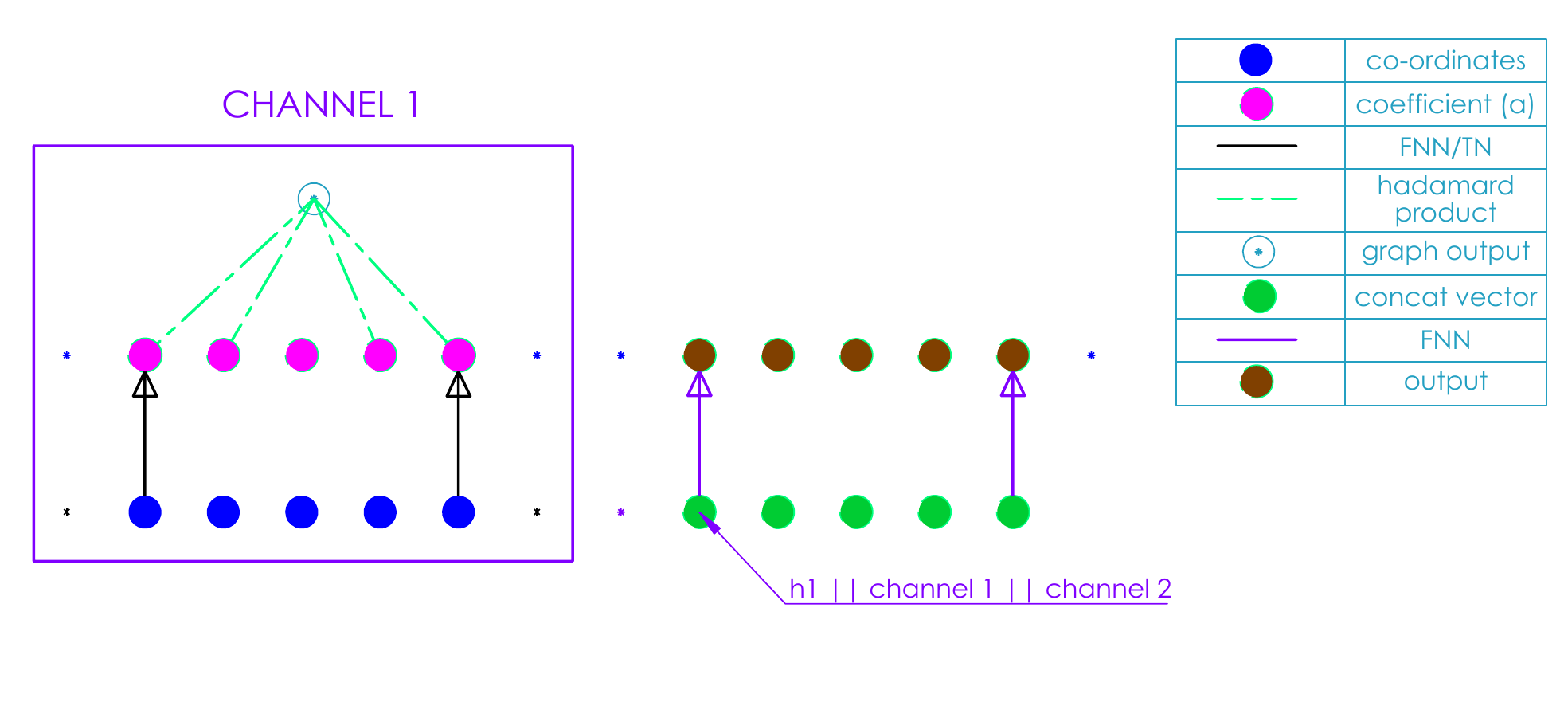}}
    \subfigure[Graph network (2D)]{\includegraphics[width=0.75\linewidth]{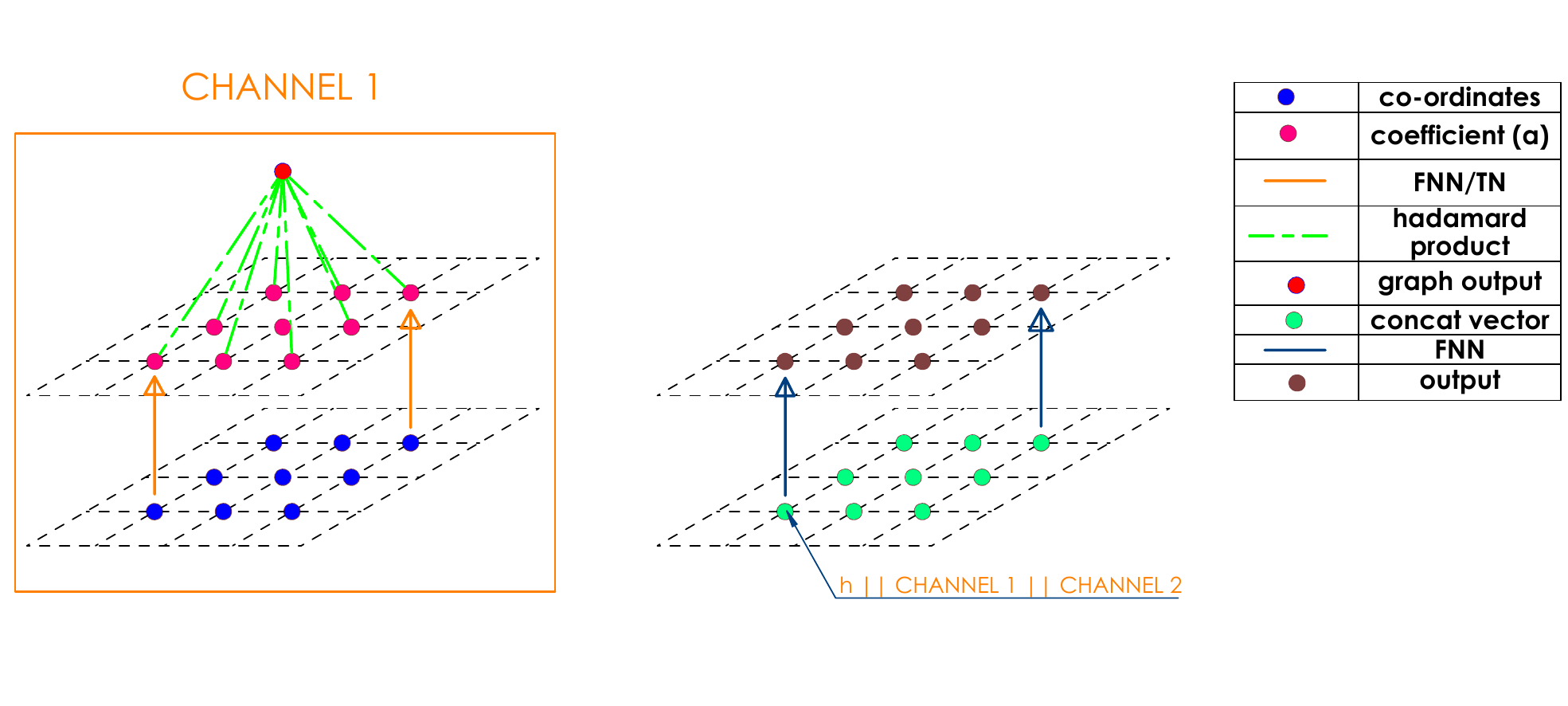}}
    \caption{Schematic representation of the proposed framework. (a) GrADE based on RK4-3/8 scheme. The magenta boxes represents $\bm y$ represented using Eq. \eqref{eq:overall_net}. (b) Proposed graph attention for 1D problem, (c) Proposed graph attention for 2D problem.}
    \label{fig:st}
\end{figure}

As the final piece of the puzzle, we concatenate the state vector $\bm u$ with the outputs of the two graph network layers, $\nabla \bm u$ and $\mathbb H \left( \bm u \right)$ and pass it through a FNN to obtain the operator $\mathcal N_N \left(\cdot; \bm \Theta \right)$ in Eq. \eqref{eq:RK4}. Mathematically, the overall operation being carried out inside the network can be represented as
\begin{equation}\label{eq:overall_net}
    \bm y = \mathcal N_N(\bm h; \bm \Theta) = \mathcal{N}_N \left(\underbrace{ \left[\bm u, \underbrace{\mathcal{GN}^{(1)}\left( \bm x, \bm u; \bm \theta_N^{(1)} \right)}_{\nabla \bm u}, \underbrace{\mathcal{GN}^{(2)} \left( \bm x, \mathcal{GN}^{(1)}\left( \bm x, \bm u; \bm \theta_N^{(1)} \right) ; \bm \theta_N^{(2)}  \right)}_{\mathbb H \left( \bm u \right)}\right]}_{\bm h}; \bm \theta_N^{(3)} \right),
\end{equation}
with $\bm \Theta = \left[ \bm \theta_N^{(1)}, \bm \theta_N^{(2)}, \bm \theta_N^{(3)} \right]$ and $\bm y = \left[\bm y_1, \bm y_2, \bm y_3, \bm y_4\right]$. The neural network is supposed to learn the dependencies of $\bm y_i$ on $\bm y_{1:i-1}$ (see Eq. \eqref{eq:RK4_pred}). Note that the graphical network includes the FNN used for inducing the attention mechanism. A schematic representation of the network architecture is shown in Fig. \ref{fig:st}. $\bm u_t$ is computed by combining Eq. \eqref{eq:overall_net} with Eq. \eqref{eq:RK4}.

\noindent \textbf{Remark 4:} We note that the number of layers used in the graph network is in accordance with the highest order of the spatial derivative. In this paper, we have limited ourselves to PDEs having second order spatial derivatives only. Also dimension of output of $\mathcal{GN}^{(1)}$ and $\mathcal{GN}^{(2)}$ is dependent upon dimension of problem, we only discuss the dimensions used for a 2D problem.

\noindent \textbf{Remark 5:} Unlike the original work on GAT in \cite{gat}, attention in GAT is introduced directly by using the nodal coordinates. This is possible because we are dealing with a physical domain where the nodal coordinates are available to us. Intuitively, GrADE assigns more ``attention'' to the edges that connects nearby nodes.

\noindent \textbf{Remark 6:} Although the mathematical expression in Eqs. \eqref{eq:H_operator} - \eqref{eq:overall_net} are expressed in matrix form, we have implemented it by expressing the same in vectorized form.

\noindent \textbf{Remark 7:} Although, we have mentioned gradient of state variables $\bm u$, $\nabla {\bm u}$ as output of the first graph network layer, this only holds when the network is trained. Similarly, once the network is trained, the second graph network layer yields the Hessian of the state variables. $\mathbb H \left( \bm u \right).$ 

\subsection{Training}
Having discussed the architecture of the proposed GrADE, we proceed to discuss the algorithm used for training the network. However, the proposed architecture blends GAT, FNN, and Neural ODE, each of which is trained differently. For instance, we generally use Message Passing (MP) algorithms for training graph networks. On the other hand, numerical integration schemes are used for training Neural ODE. Therefore, training GrADE naturally involves both MP and numerical integration, with MP being used in the spatial domain and numerical integration being used in the temporal domain.

Consider $v_i$ to be the $i-$th node in the graph network. In MP, we update the state of $v_i$ based on information from its neighbors $N_i$. The MP step can be divided into two steps: message aggregation and state update. In the message aggregation step, the messages received from all the nodes are aggregated into a single message. In the first graph layer of GrADE, the message aggregation step represents computing $\bm H \left(\gamma_{i,j},\delta \bm_{i,j}\right)$ in Eq. \eqref{eq:H_operator} for each neighboring nodes followed by mean operation. Similarly, for the second graph layer of GrADE, the message aggregation step represents computing the mean in Eq. \eqref{eq:cust_graph2}. The update step involve computing $\nabla \bm u$ from $\bm u$ in the first graph layer, and $\mathbb H \left( \bm u \right)$ from $\nabla \bm u$ in the second graph layer. Once the updates for $\nabla \bm u$ and $\mathbb H \left( \bm u \right)$ are available, we utilize the same in computation of the numerical integration. As already stated earlier, RK4 scheme is used in this paper for numerical integration. 

 One major advantage of the proposed GrADE resides in the fact that time is not an explicit variable in the proposed framework; this allows the generalize the model better to future time-step. During the training phase, we allow the model to gradually explore the system and learn the network parameters. During the initial epochs, the proposed GrADE only explores a few steps starting from the initial condition. Slowly, as the model starts to learn, we allow GrADE to explore further time-steps. In practice, this is achieved by introducing a list variable $\tau_l$ that stores the number of time-steps GrADE is suppose to explore during each epoch. With this setup, GrADE is able to learn the dynamics that may be significantly different from the initial conditions and it neighbors. We also allow the learning rate $\eta$ to vary with epoch by maintaining another list variable $\eta_l$. Overall, we implemented RK4 schme using the open-source library \texttt{torchdiffeq} \cite{Node}. For ease of understanding, an algorithm depicting the training procedure is shown in Algorithm \ref{alg:train}. 


\begin{algorithm}[ht!]
\caption{Training GrADE}
\label{alg:train}

\textbf{Inputs:} $\mathcal D = \left\{\bm x, t_i, \bm U\left( \bm x, t_i \right) \right\}_{i=1}^{N_s}$. \\

\textbf{Set Hyperparameter:} Number of epochs $N_e$, time-step $\Delta t$, $\tau_l$, and $\eta_l$. \\


\textbf{Initialize:} Neural network model: $\mathcal N_N(\cdot; \bm \Theta)$\Comment*[r]{Eq. \eqref{eq:overall_net}}

\For{epoch $= 0$ \textbf{to} $N_e$}{
    $\tau \leftarrow \tau_l[epoch]$
    
    $\eta \leftarrow \eta_l[epoch]$ 
    
    Formulate $\bm U_t$ using $\bm U\left( \bm x, t_i \right)$ and $\tau_l[epoch]$\Comment*[r]{Target for current epoch}
    
    \For{t $= 0$ \textbf{to} $\tau$}{
        $\bm u^{t+1} \leftarrow \bm u^{t} + \Delta t \times \bm {\mathcal F}\left( \mathcal N_N(\cdot; \bm \Theta) \right)$\Comment*[r]{Combination of Eqs. \eqref{eq:overall_net} and \eqref{eq:RK4_pred}} 
        
        $\bm U_p[t] \leftarrow \bm u^{t+1}$\Comment*[r]{Store GrADE prediction} 
        }
        
    $\mathcal L = MSE(\bm U_p, \bm U_t)$\Comment*[r]{Calculate loss}
    $\bm \nabla \bm w \leftarrow$ Backprop($\mathcal L$) \\
    $\bm w \leftarrow \bm w - \eta \bm \nabla \bm w$ \Comment*[r]{Update weights}
    } 

\textbf{Output:} Trained model $\mathcal N_N(\cdot; \bm \Theta)$.

\end{algorithm}

\section{Numerical implementation and results}\label{sec:ne}
We consider the well-known Burgers' equation for illustrating the performance of the proposed GrADE. Burgers' equation is a fundamental PDE occurring in various areas of applied mathematics, such as fluid mechanics, nonlinear acoustics, gas dynamics, and traffic flow. We solve Burger' equation in both 1D and 2D. For both cases, the simulation data is generated by using open-source FE solver, FeNICS \cite{alnaes2015fenics}.



\subsection{1D viscous Burgers’ equation}

First, we consider 1D viscous Burgers' equation with periodic boundary  

\begin{equation}\label{eq:burger1D}
    \frac{\partial u}{\partial t} + u \frac{\partial u}{\partial x} - \nu \frac{\partial^2 u}{\partial x^2} = 0
\end{equation}
\begin{equation}\label{eq:Burger1D_bc}
    u(x=0, t) = u(x=L, t), \ x \in [0, L], \ t \in [0, T],
\end{equation}
where $u$ is the velocity and $\nu = 0.0025$ in viscosity. We consider random initial condition given by a Fourier series with random coefficients
\begin{equation}\label{eq:IC}
    u(x, t=0) = \frac{2w(x)}{\max_x|w(x)| + c},
\end{equation}
where
\begin{equation}\label{eq:burger1D_wx}
    w(x) = a_0 + \sum_{l=1}^{N_l} a_l \sin (2l\pi x) + b_l\cos(2l\pi x).
\end{equation}
In Eq. \eqref{eq:burger1D_wx}, $a_l, b_l \sim N\left(0,1\right)$ are drawn from standard Gaussian distribution and $c \sim \mathcal U \left(-1,1\right)$ is drawn from a uniform distribution. We have considered $L=1$ in Eq. \eqref{eq:Burger1D_bc} and $N_l = 4$ in Eq. \eqref{eq:burger1D_wx}.

For generating data using FeNICS, we discretized the spatial domain into 512 points and use a time-step $\Delta t=0.001$. For training and testing the proposed GrADE, we use $\Delta t = 0.007$. All the three FNNs present within the proposed GrADE are considered to be shallow nets with only hidden layer. We use LeakyReLU activation functions with a negative slope of 0.2 for all FNNs. For the two attention nets $\mathcal N_{N_1}\left(\cdot; \bm \theta_N^{(1)}\right)$ and $\mathcal N_{N_2}\left(\cdot; \bm \theta_N^{(2)}\right)$, the hidden layer has 32 neurons. As for the third network $\mathcal N_N\left(\cdot;\bm \theta_N^{(3)}\right)$, the hidden layer has 32 neurons. As we are dealing with a 1D problem here, all the three FNNs have only one output each. Overall the proposed GrADE has 387 parameters. 

\begin{figure}[hbt!]
    \centering
    
    \subfigure[]{\includegraphics[width=0.46\linewidth]{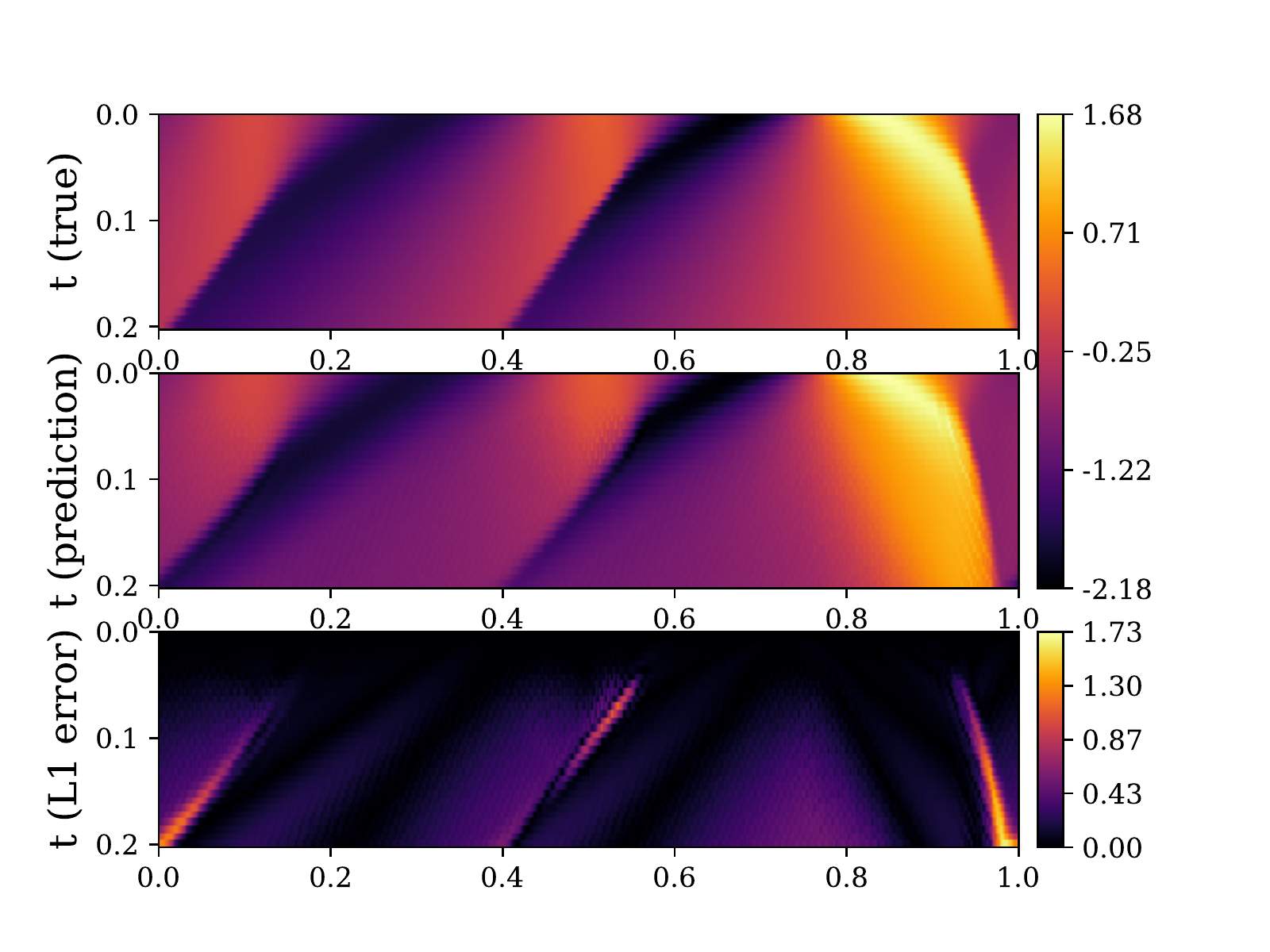}
    \label{fig:trs_bt_test1}}
    \subfigure[]{\includegraphics[width=0.46\linewidth]{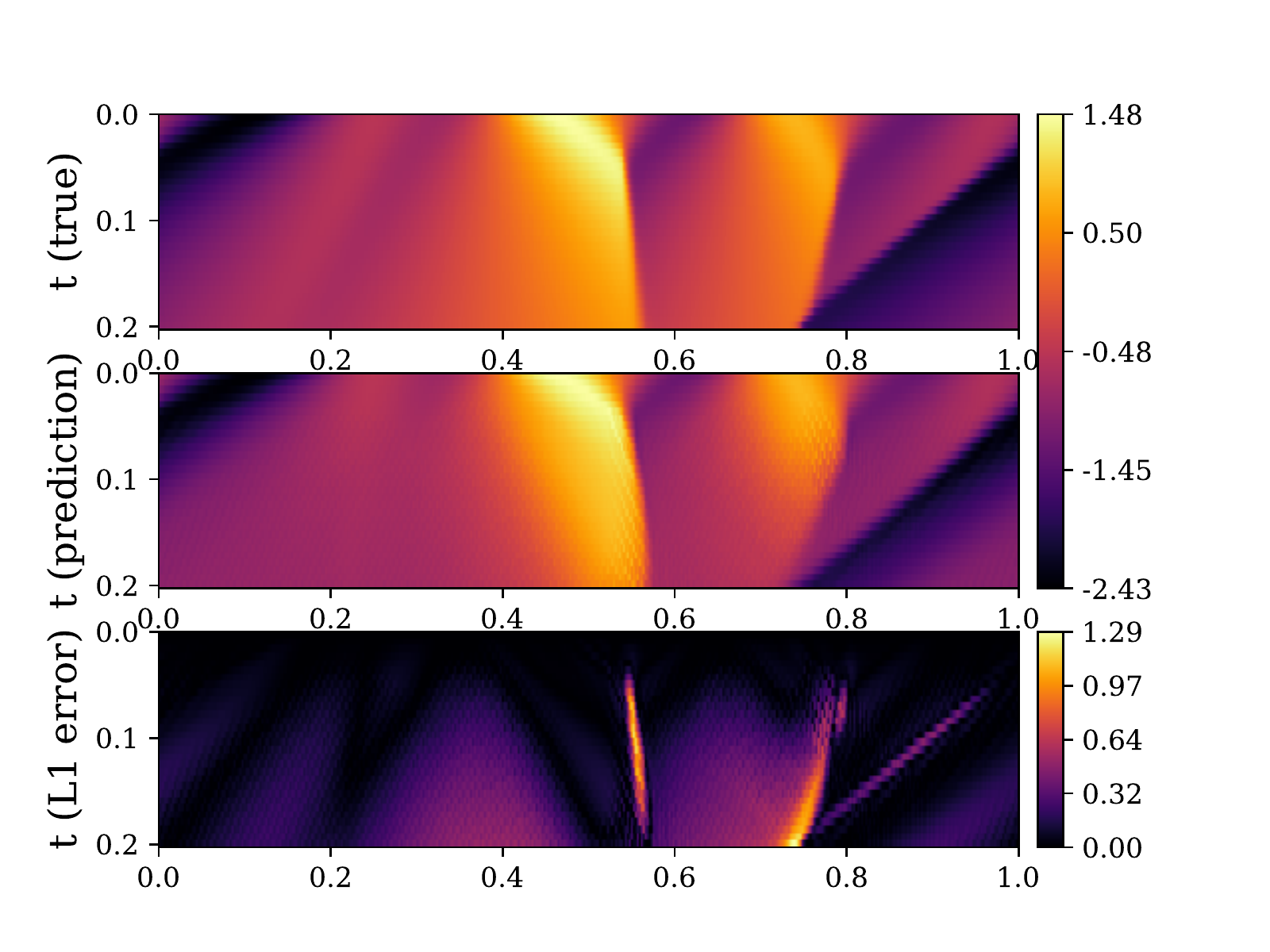}
    \label{fig:trs_bt_test2}}
    
    \caption{Figure depicting evolution of predicted velocities of 1D Burgers' equation with two initial conditions. First, second, third row shows FEM simulation, network prediction and L1 error, respectively.}
    \label{fig:1D_sample}
\end{figure}

We trained the proposed GrADE using 120 samples of the initial condition and snapshots at four time instants only (i.e., last integration time index is 4). We use a learning rate of 0.07 and train the model for 201 epochs. For testing, we used 30 additional realizations of the random initial conditions.
Fig. \ref{fig:1D_sample} shows the solutions of 1D Burgers' equation for two random initial condition (from the test set) obtained using FeNICS (first row) and GrADE (second row). While $x-$ axis in Fig. \ref{fig:1D_sample} represents the spatial domain,  $y-$ axis represents the temporal domain. The third row in Fig. \ref{fig:1D_sample} represents the L1 error between the FeNICS and the GrADE results. We observe that results obtained using GrADE and FeNICS matches almost exactly. We note that the underlying dynamics is extremely complex due to formation of shocks. It is impressive that the proposed model trained with data at four temporal snapshots only (with $\Delta t - 0.007)$ is able capture the temporal evolution of the system dynamics far beyond the training regime (up to $0.2$s).

\begin{table}[!hbt]
    \centering
    \caption{Hyper-parameters of GrADE for Burgers' 1D equation. * represents number of items in list, similar to python list notation.}
    
    \label{tab:exp1}
    \begin{tabular}{p{2cm}p{9cm}p{2cm}p{3cm}}
        \hline 
            \textbf{lit train (index)} & \textbf{learning rate list} & \textbf{training epochs} & \textbf{Training scenarios} \\
        \hline
            2 & [0.07]*201 & 201 & 120\\
            3 & [0.07]*401 & 401 & 120\\
            4 & [0.05]*25 + [0.052]*25 + [0.054]*50 + [0.056]*301 & 401 & 120\\
            5 & [0.045]*25 + [0.048]*25 + [0.052]*50 + [0.054]*301 & 401  & 120\\
        \hline
    \end{tabular}
\end{table}

To illustrate the robustness of the proposed GrADE, we perform numerical experiments by varying the last integration time index and number of training scenarios provided to GrADE during training. For efficient training, different network hyperparameters have been used for different case. Details on the same is provided in Table \ref{tab:exp1}. For comparing the accuracy in prediction, we compute the L2 error between the GrADE predicted results and true solution at each time-index as follows:
\begin{equation}\label{eq:odeint}
    \epsilon_j = \sum_{i=1}^{N_{s}}|| u_{p,j} - u_{t,j}||_2^2,
\end{equation}
where $N_s$ denotes the number of test samples, $u_{p,j}$ is the Grade predicted result at time-index $j$, and $u_{t,j}$ represents the target obtained using FeNICS. $\epsilon_j$ is the error at  time-index $j$.

\begin{figure}[hbt!]
    \centering
    \subfigure[]{\includegraphics[width=0.4\linewidth]{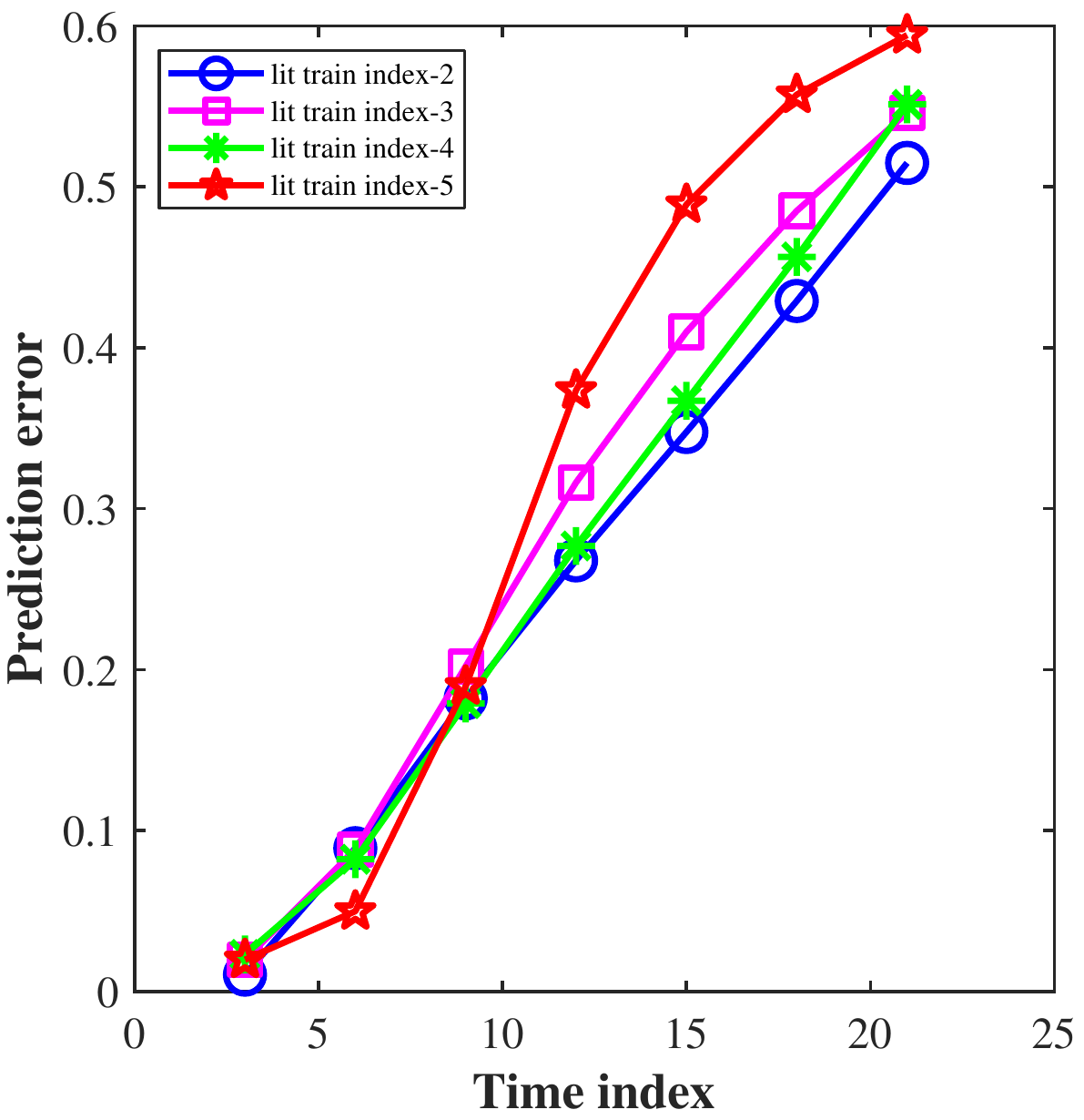}
    \label{fig:lit_train__lit_test_1d}}
    \subfigure[]{\includegraphics[width=0.4\linewidth]{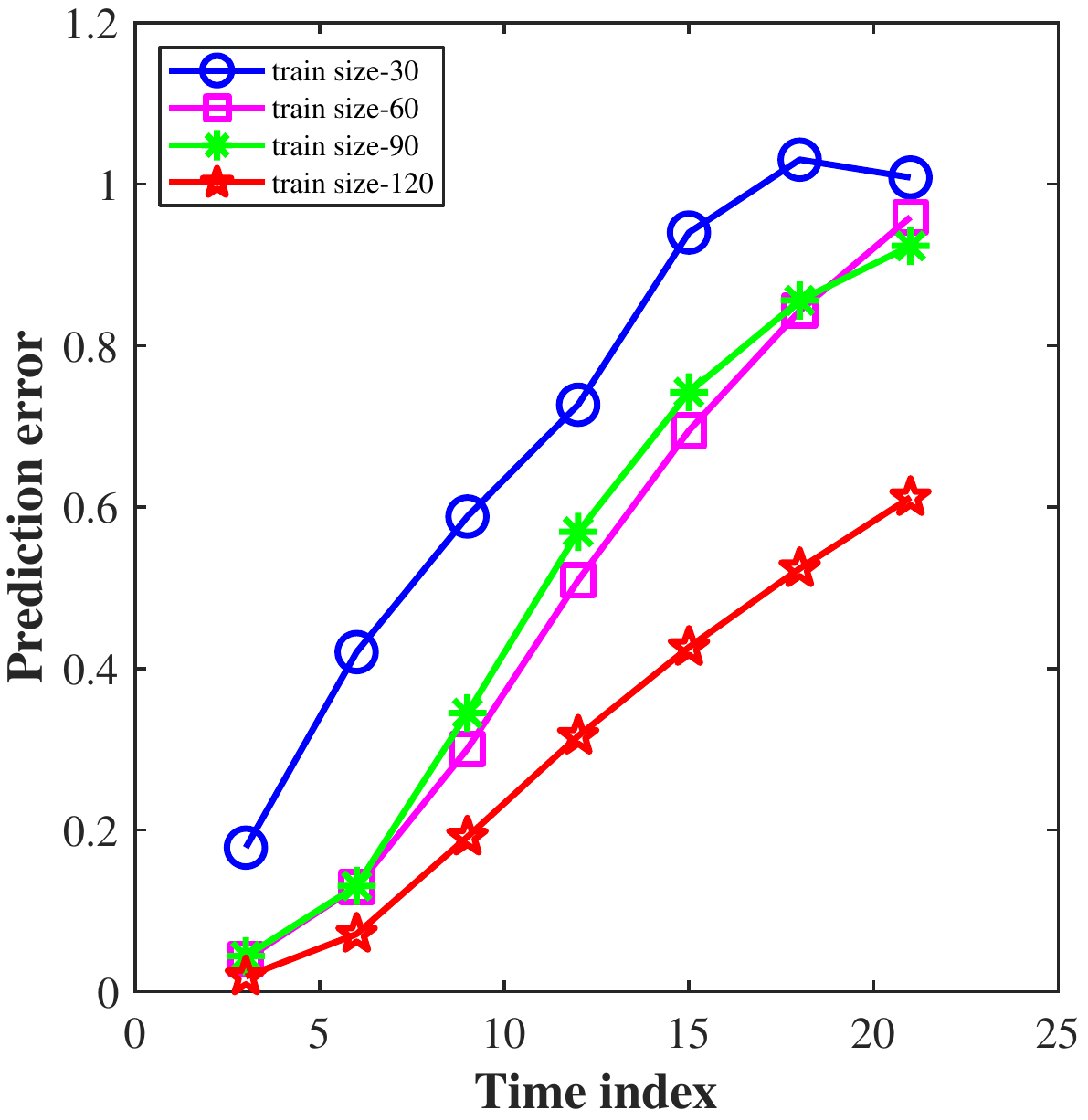}
    \label{fig:tr_size__lit_test_1d}}
    
    \caption{Burgers' 1d: (a) Prediction error with time for models trained with different last integration time index (LIT), where $N_t$ is max time used during training (b) Prediction error with time for model trained with different number of training graph. Note that actual time is $Time \: index \times \Delta t$, where $\Delta t=0.007$}
    \label{fig:1D}
\end{figure}

Fig. \ref{fig:lit_train__lit_test_1d} shows results for Experiment 1 which compares prediction error with increasing time, for networks trained on different last integration time index. We notice that for last integration time index of two, three and four, the network has identical predictive capability. However, for last integration time index of 5, the result starts deviating beyond time-index 5, indicating over-fitting. As for computational time, the network trained with higher last integration time index takes more time to train. 
Fig \ref{fig:tr_size__lit_test_1d} shows results for Experiment 2 which compares prediction error with increasing time for networks trained with different number of training graphs (training scenarios). We use a learning rate of $0.07$ and last integration time index of 4 for all training sizes. As expected, we observe that the best result is obtained with 120 training scenarios and the worst with 30 scenarios. Results obtained with 60 and 90 scenarios are almost same.

\begin{figure}[hbt!]
    \centering
    \includegraphics[width=0.6\linewidth]{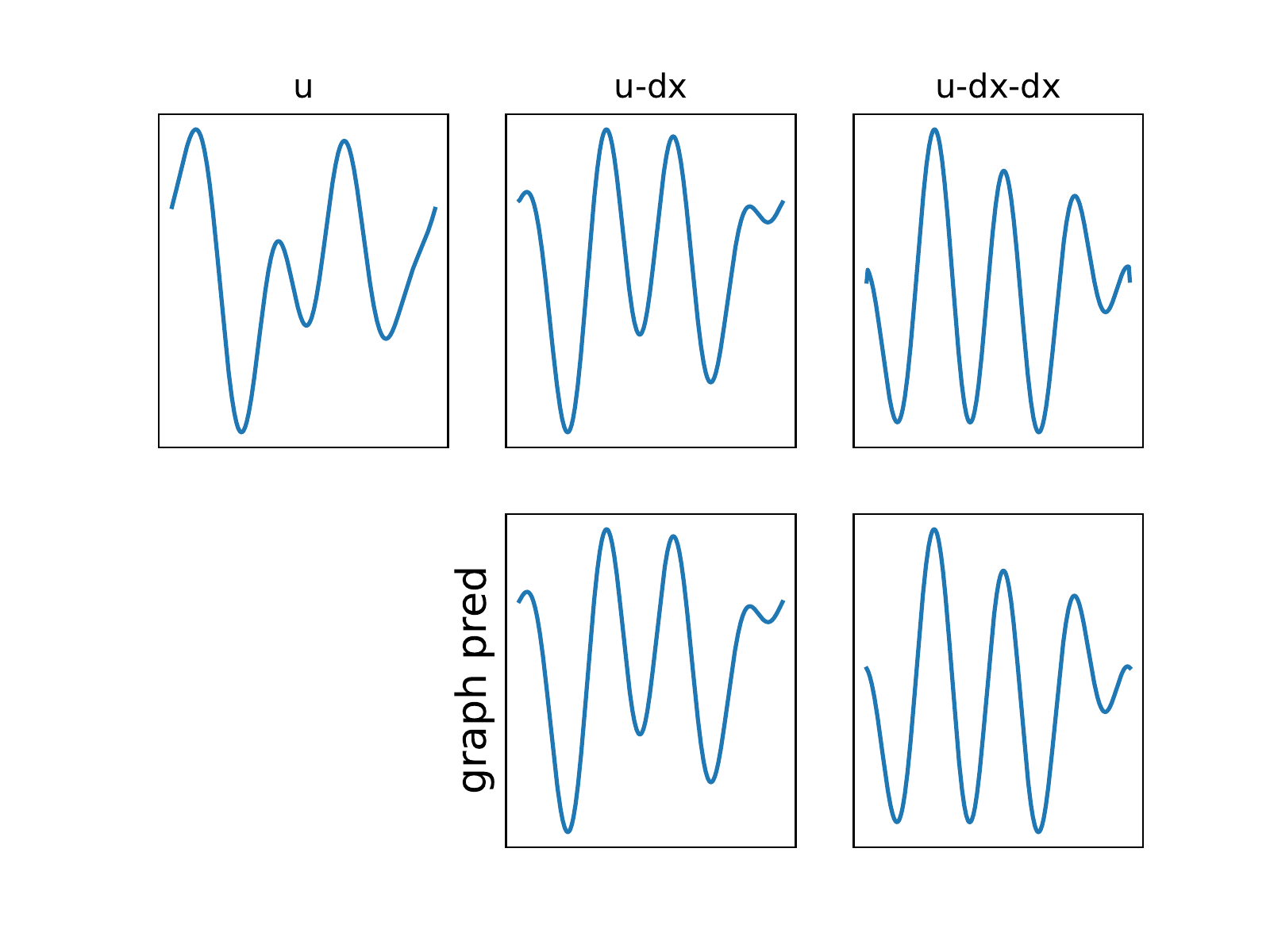}
    
    \caption{Caparison of output of two graph network layers used in model and derivatives of data computed using central differences.}
    \label{fig:graph_out1d}
\end{figure}

Finally, we examine the output of the GAT present within the proposed GrADE. As stated earlier, once trained, the output of the first and second graph network layers should yield spatial derivatives $u_x$ and $u_{xx}$, respectively. In Fig \ref{fig:graph_out1d}, we compare the outputs of the two graph network layers with the derivative obtained using central difference scheme. Excellent match between the two is observed.

\subsection{2D coupled Burgers’ equation}

\begin{figure}[hbt!]
    \centering
    \includegraphics[width=1\linewidth]{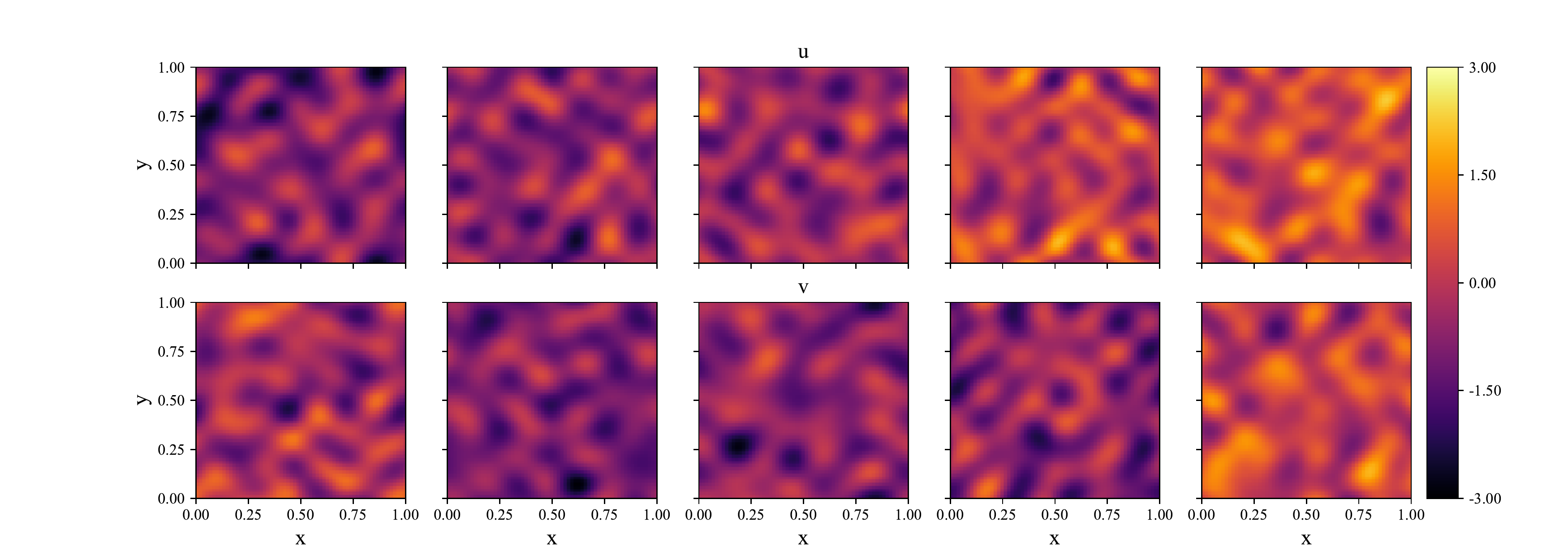}
    
    \caption{Randomly generated initial condition for x and y velocity-components using truncated Fourier series.}
    \label{fig:ic}
\end{figure}

We consider the the 2D coupled Burgers' system. It has the same convective and diffusion form as the incompressible Navier-Stokes equations. It is an important model for understanding of various physical flows and problems, such as hydrodynamic turbulence, shock wave theory, wave processes in thermo-elastic medium, vorticity transport, dispersion in porous medium. Numerical solution of Burgers’ equation is primary step towards when developing methods for complex flows. The governing equations for Burgers' equation takes the following form:
\begin{equation}
\label{eq:b2}
\bm u_t +\bm u \cdot \nabla \bm u - \nu \Delta \bm u = 0,
\end{equation}
with periodic boundary condition
\begin{equation}\label{eq:bc_b2}
\begin{split}
    \bm u \left(x=0, y, t \right) & =\bm u \left(x=L, y, t \right), \\
    \bm u \left(x, y=0, t \right) & =\bm u \left(x, y=L, t \right).
\end{split}
\end{equation}
Eq. \eqref{eq:b2} can be written in expanded form as
\begin{equation}
\label{eq:b2m}
\begin{aligned}
    \frac{\partial u}{\partial t} + u \frac{\partial u}{\partial x} + v \frac{\partial u}{\partial y} - \nu (\frac{\partial^2 u}{\partial x^2} + \frac{\partial^2 u}{\partial y^2}) = 0  \\
    \frac{\partial v}{\partial t} + u \frac{\partial v}{\partial x} + v \frac{\partial v}{\partial y} - \nu (\frac{\partial^2 v}{\partial x^2} + \frac{\partial^2 v}{\partial y^2}) = 0,
\end{aligned}
\end{equation}
where $\nu=0.005$ is viscosity, $u$ and $v$ are the $x$ and $y$ components of velocity. We consider $\{x,y\} \in [0,1]$.
Similar to the 1D case, the initial condition is defined using truncated Fourier series with random coefficients:
\begin{equation}\label{eq:ic2}
   \bm  u(x, y, t=0) = \frac{2 \bm w(x, y)}{\max_{\{x, y\}} |\bm w(x, y)|} + \bm c,
\end{equation}
where
\begin{equation}
   \bm w(x, y) = \sum_{i=-L}^{N_l}\sum_{j=-L}^{L} \bm a_{ij} \sin(2 \pi (ix+jy)) + \bm b_{ij} \cos(2 \pi(ix+jy)),
\end{equation}
where $\bm a_{ij}, \bm b_{ij} \sim \bm N(0, \mathbf I_2)$, $L=4$ and $\bm c \sim \bm {\mathcal U}(-1, 1) \in \mathbb{R}^2$. Some representative initial conditions generated using Eq. \eqref{eq:ic2} are shown in Fig. \ref{fig:ic}.

Similar to the 1D case, we use FeNICS to generate the training data. We discretize the spatial domain in FeNICS into $64 \times 64$ grid and use a time-step of $0.005$. For the FNNs present within the proposed GrADE, we consider shallow nets with only one hidden layer. For the two attention nets $\mathcal N_{N_1}\left(\cdot;\bm \theta_N^{(1)} \right)$ and $\mathcal N_{N_2}\left(\cdot;\bm \theta_N^{(2)} \right)$, the hidden layer has 32 neurons. The hidden layer of the third network $\mathcal N_N \left(\cdot; \bm \theta_N^{(3)} \right)$ has 64 neurons. Overall, the proposed GrADE has 2446 trainable parameters.

\begin{figure}[hbt!]
    \centering
    \subfigure[]{\includegraphics[width=0.46\linewidth]{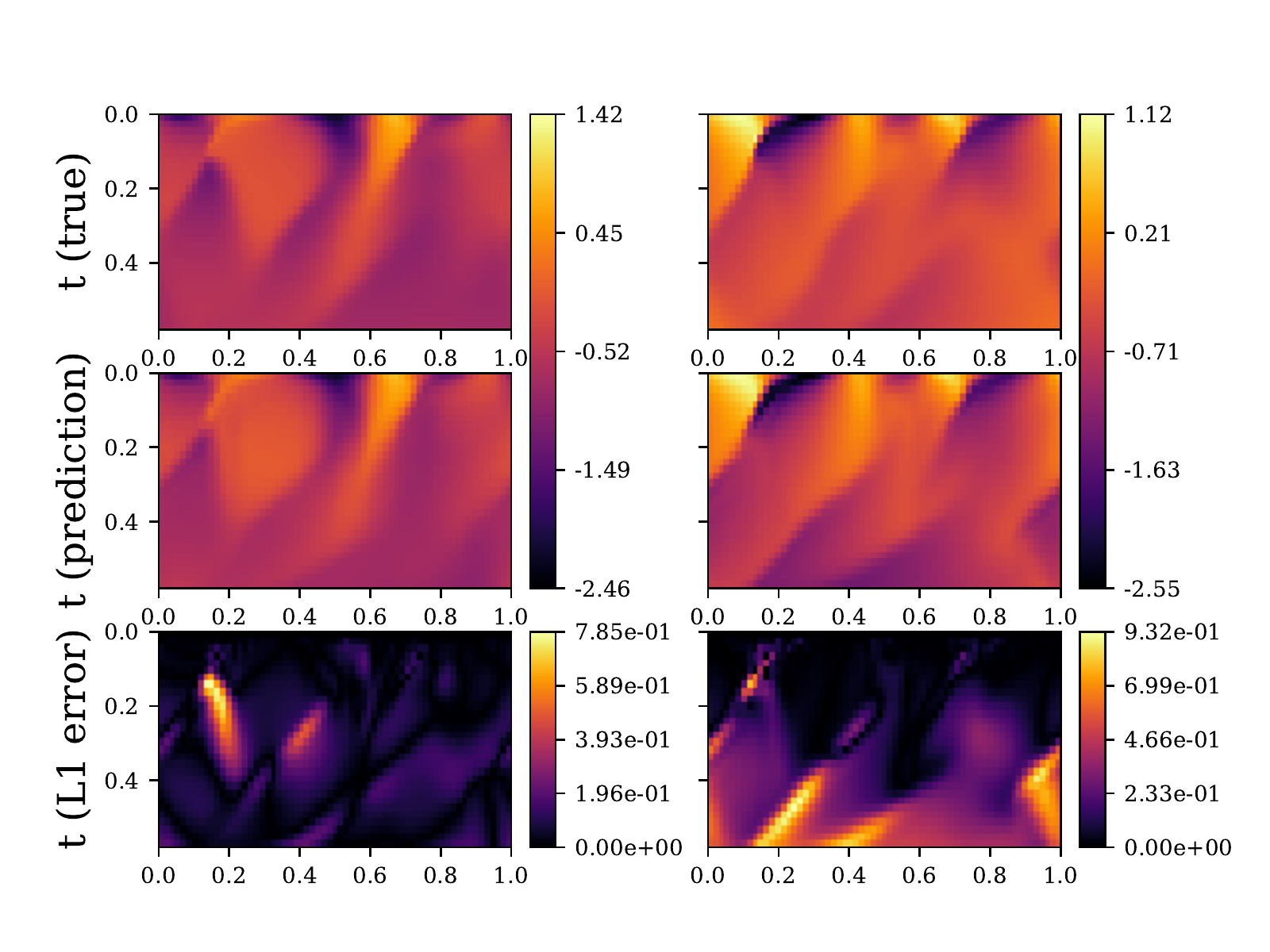}
    }
    \subfigure[]{\includegraphics[width=0.46\linewidth]{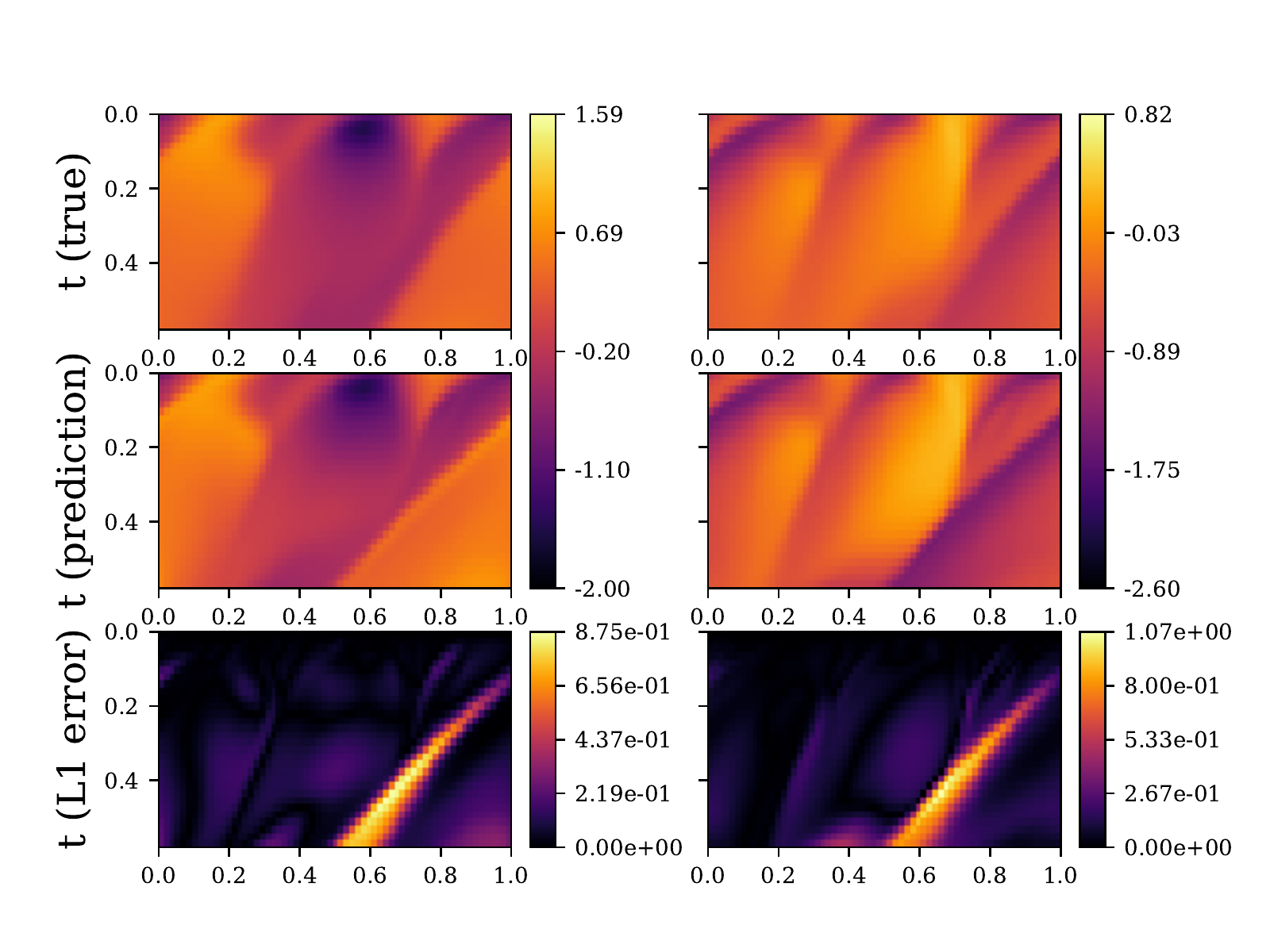}
    }
    \caption{Figure depicting domain slice evolution of predicted x and y velocities of 2D coupled Burgers' equation with two initial conditions. Last row depicts the L1 error.}
    \label{fig:B2D_slice_pred}
\end{figure}

\begin{figure}[htbp!]
    \centering
    \includegraphics[width=1\linewidth]{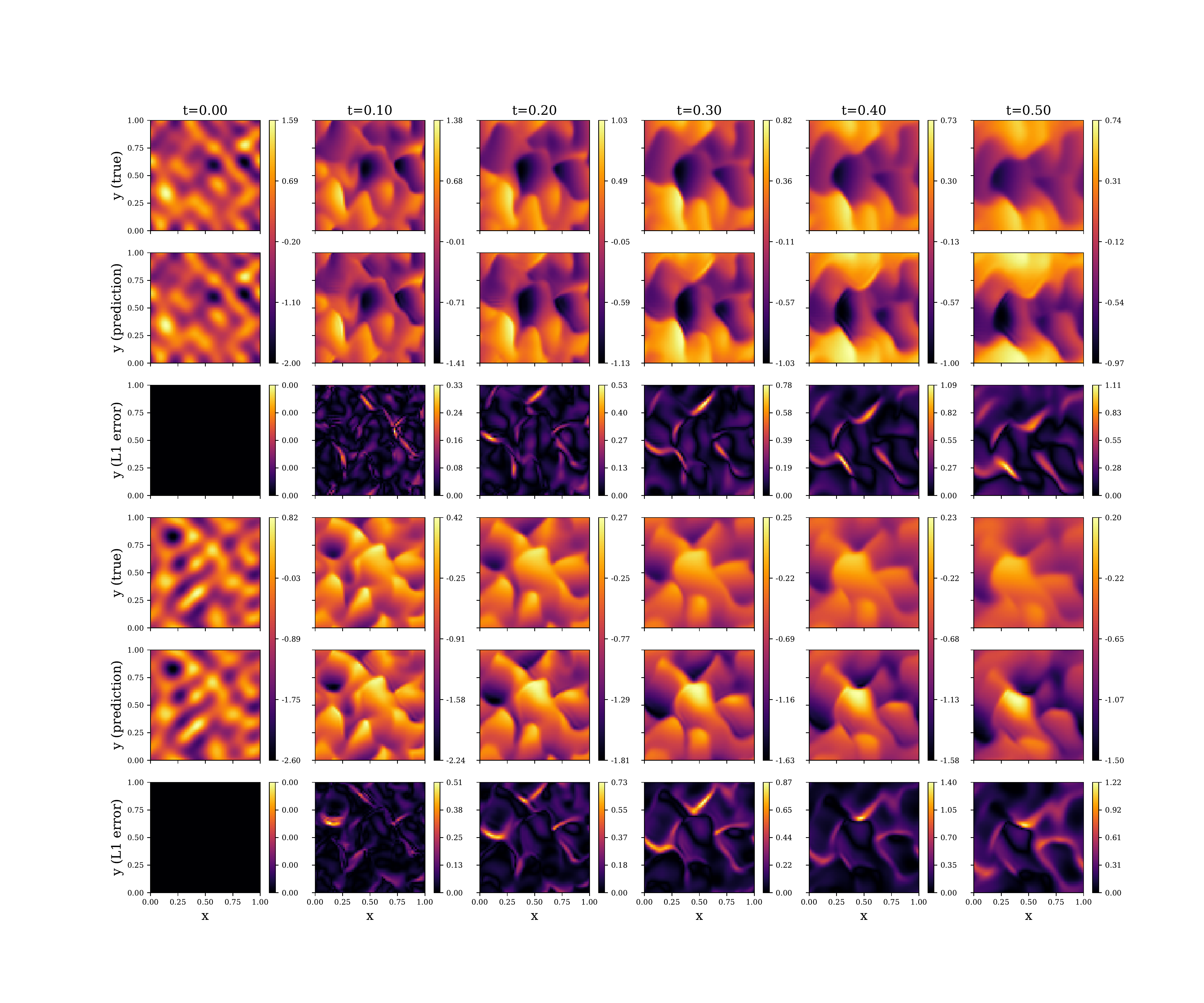}
    \caption{Prediction of x and y velocities of 2D coupled Burgers' equation at different time steps. Top to bottom, three rows shows x-velocity, y-velocity, L1 error related plots, respectively.}
    \label{fig:B2D_pred}
\end{figure}

We trained the proposed GrADE using 120 samples of the initial condition and snapshots at three time instants only (i.e., last integration time index is 3). We allowed the learning rate to vary with number of epochs and trained the model for 501 epochs. For testing, we generated 20 additional realizations of the random initial conditions. Fig. \ref{fig:B2D_slice_pred} shows the the results corresponding to two initial conditions from the test dataset obtained using FeNICS
and the proposed GrADE. The first and second rows depict the velocities (slices along $x$ and $y$ axes) obtained using FeNICS and the proposed approach respectively. Reasonable match among the results is observed. The third column shows the L1 error. Fig. \ref{fig:B2D_pred} shows the velocities at different time-steps obtained using FeNICS and the proposed approach. Note that predicting the velocities for the 2D case as well is extremely difficult because of the formation of shocks. The fact that the proposed GrADE trained with observations at only three snapshots (with $\Delta t = 0.02$) is able to provide reasonably accurate results is really impressive.

\begin{figure}
    \centering
    \includegraphics[width=0.5\textwidth]{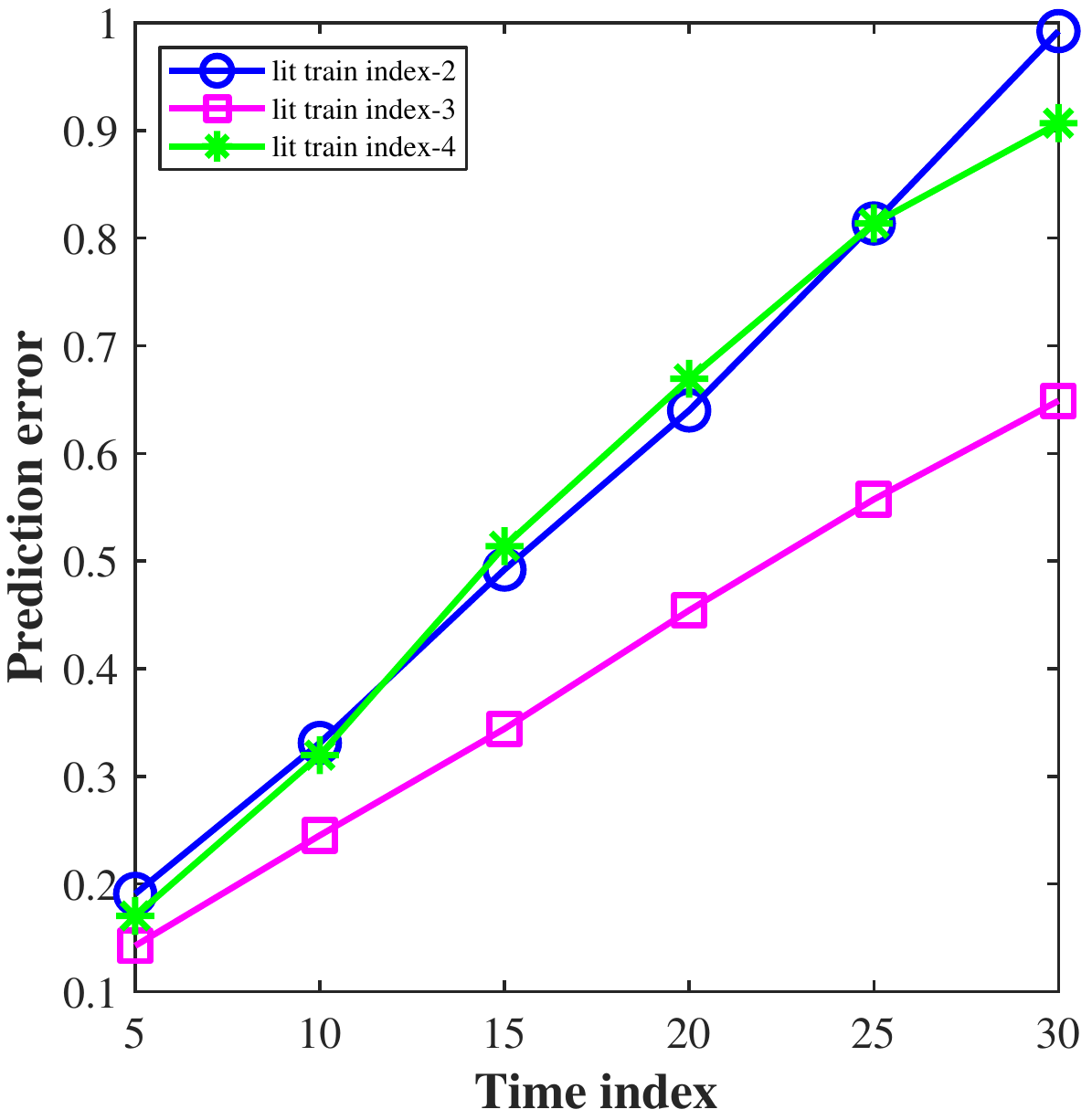}
    \caption{Prediction error with increasing time for model trained with different number of training graph}
    \label{fig:lit_train__lit_test_2d}
\end{figure}

\begin{figure}
    \centering
    \includegraphics[width=0.5\textwidth]{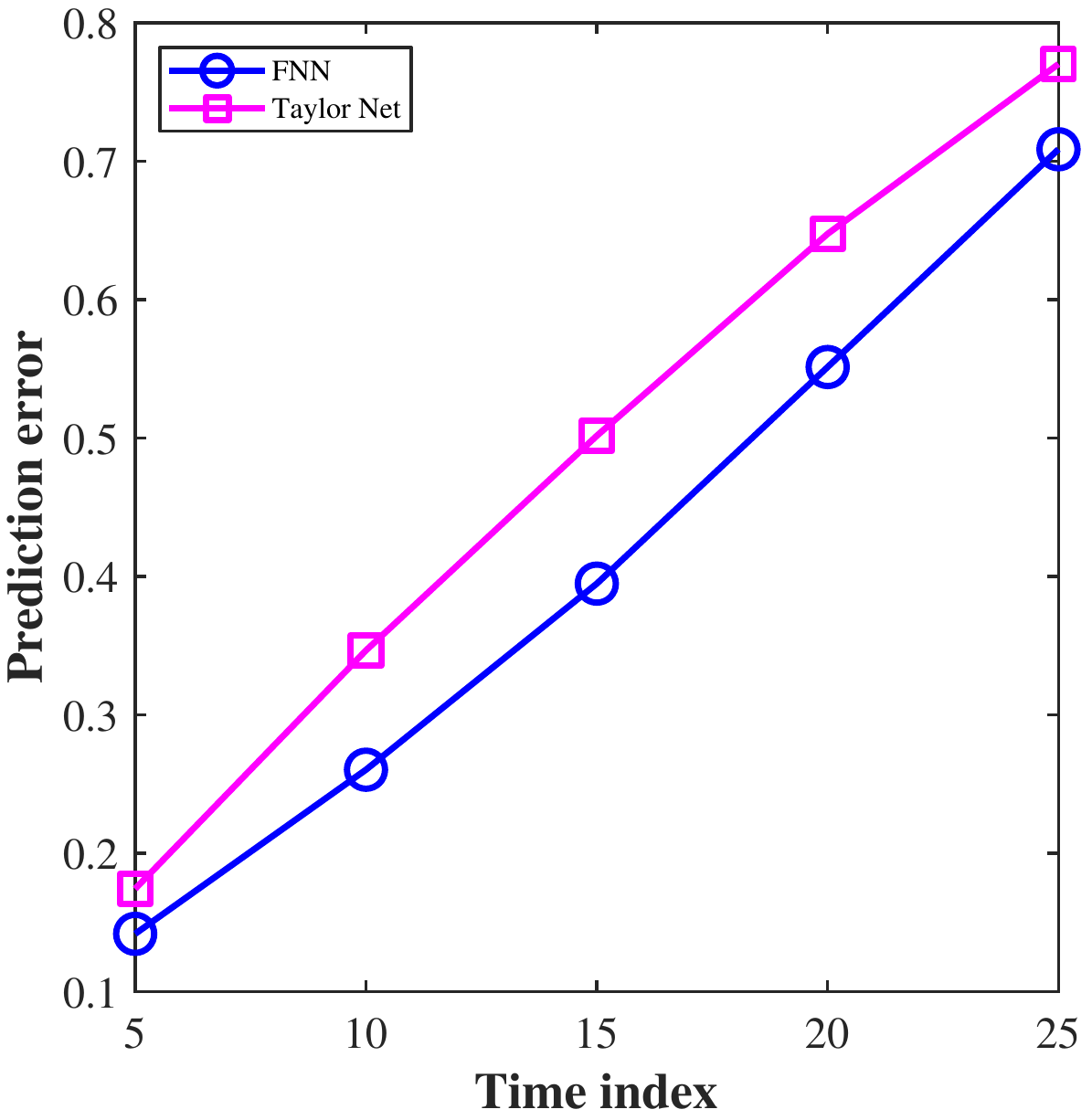}
    \caption{Prediction error with time for model using a FNN and Taylor net for $\mathcal N_{N_1}$ and $\mathcal N_{N_2}$ in eq. \ref{eq:cust_GAT} and \ref{eq:NN2}}
    \label{fig:layer__lit_test_2d}
\end{figure}

To understand the influence of last integration time index used during training, we perform case study by varying the last integration time index. The hyperparameter setting for all the cases are shown in Table \ref{tab:exp1_B2D}. The results obtained are shown in Fig. \ref{fig:lit_train__lit_test_2d}. Unlike the 1D Burgers' equation where the error was almost similar for all the cases, error is least when trained with last integration time index of 3. This is probably because the loss-function becomes extremely complex on increasing the last integration time index during training beyond 3. One way to address this issue is to use depth refinement. The idea is to alter depth of network during training. We use a smaller last integration time index during the initial training; however, as the training progresses, we start increasing the depth of the model. We recall that depth in GrADE refers to the number of time integration steps. This method enable us to train networks with more depth. The results corresponding to depth refinement are shown in Fig. \ref{fig:layer__lit_test_2d}. We start with a depth of 2 and gradually increased it till depth 4. Note that extra care is necessary with the depth refinement framework. The hyperparameters used are shown in Table \ref{tab:exp1_B2a}. Results are compared with those obtained using a constant depth of 4. We observe that the computational time needed is less and the accuracy of the model is better for the depth refinement framework.

\begin{table}[!hbt]
    \centering
    \caption{Hyper-parameters of GrADE for Burgers' 2D equation for Experiment 2. * represents number of items in list, similar to python list notation.}
    
    \label{tab:exp1_B2D}
    \begin{tabular}{|p{2cm}|p{9cm}|p{2cm}|p{3cm}|}
        \hline 
            lit train & learning rate list at each training epoch & training epochs & training initial conditions \\
        \hline
            2 & [0.055]*200 + [0.053]*100 + [0.05]*101 & 401 & 90\\
        \hline
            3 & [0.055]*200 + [0.054]*100 + [0.03]*101 & 401 & 90\\
        \hline
            4 & [0.045]*400 + [0.044]*301 & 701 & 90\\
        \hline
    \end{tabular}
\end{table}

\begin{table}[htbp!]
    \centering
    \caption{Hyper-parameters of GrADE for Burgers' 2D equation for Experiment 3. * represents number of items in list, similar to python list notation. Model 1 is constant depth and model 2 is with depth refinement}
    \label{tab:exp1_B2a}
    \begin{tabular}{|p{1cm}|p{2cm}|p{8cm}|p{2cm}|p{3cm}|}
        \hline 
            model type & lit train & learning rate list at each training epoch & training epochs & training initial conditions \\
        \hline
            1 & 4 & [0.045]*701 & 701 & 90\\
        \hline
             2 & [2] * 200 + [3] * 300 + [4] * 201 & [0.06] * 200 + [0.022] * 25 + [0.024] * 25 + [0.032] * 50 + [0.04] * 200 + [0.015] * 25 + [0.018] * 25 + [0.022] * 25 + [0.032] * 25 + [0.04] * 101 & 701 & 90\\
        \hline
    \end{tabular}
\end{table}

\begin{figure}
    \centering
    \includegraphics[width=0.5\textwidth]{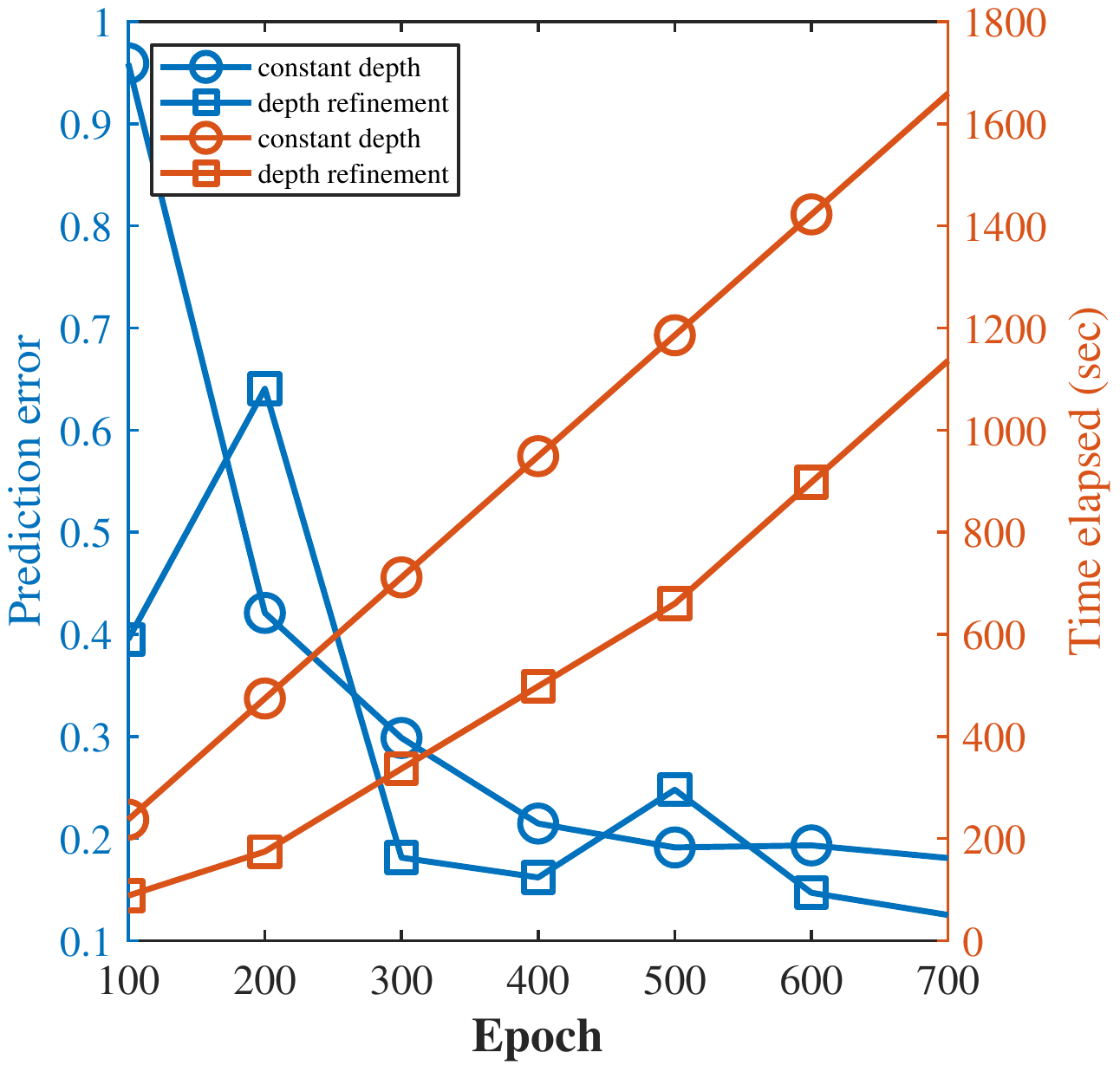}
    \caption{Plot depicts prediction error and time elapsed at different training epochs during training GrADE with constant depth versus with depth refinement.}
    \label{fig:depth_refinement}
\end{figure}

\begin{figure}[hbt!]
    \centering
    \includegraphics[width=0.8\textwidth]{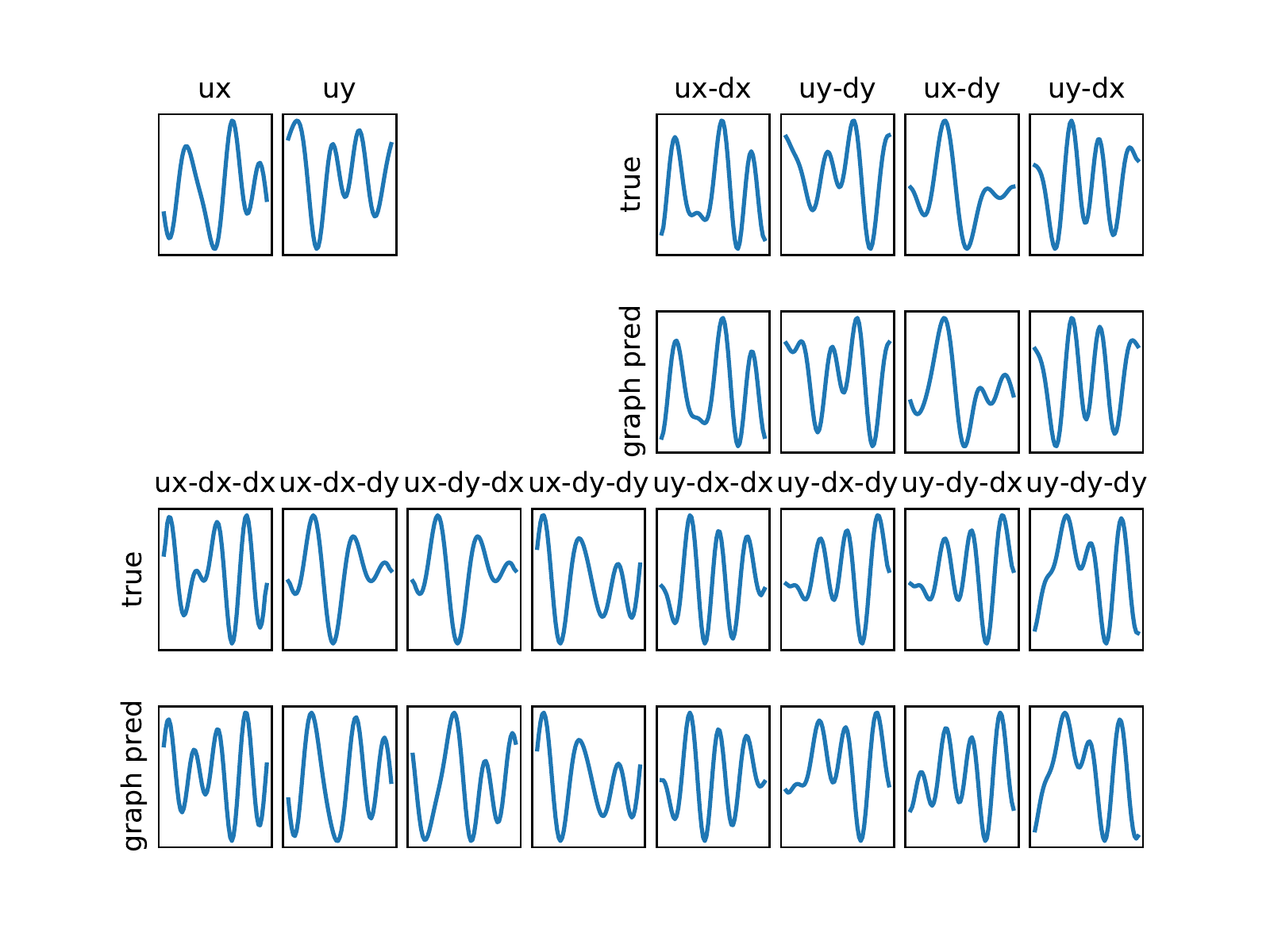}
    \caption{Caparison of output of two graph network used in model and derivatives of data computed using central differences.}
    \label{fig:graph_out}
\end{figure}

Next we concentrate on the role of attention model within the proposed framework. In this work, we have used FNN for computing the attention weights $\bm \beta$ and $\bm \gamma$. An alternative to this is to use Taylor net. SpiderConv proposed in \cite{spidercnn} uses Taylor net in GNN for classification and segmentation tasks.
It can be formulated as
\begin{equation}
\label{eq:taylor_net}
\begin{aligned}
    \bm \gamma_{i,j} = \sum_{k=0}^{Q} w_k * p_k(\bm \delta \bm x_{i,j}), \\ 
\end{aligned}
\end{equation}
where $p_k$ is element of $p \in \pi_m(\mathbb{R}^2)$, $m$ is degree of polynomials, $Q$ is number of monomials and $w_k$ are trainable weights of networks. 
In Fig. \ref{fig:layer__lit_test_2d}, we present a comparative assessment between model accuracy when using FNN and Taylor net. We use $m=3, Q=10$ in Eq. \eqref{eq:taylor_net}. We observe that the results obtained using the FNN is slightly more accurate as compared to Taylor net. Recall that the output of the first graph network layer is supposed to yield the first derivative and those of the second layer is supposed to yield the second derivative. To validate the same, we plot the output of the two graph network layers in Fig. \ref{fig:graph_out}. Results obtained using central difference are also shown. A reasonably good match between the output of the graph network layers and those obtained using central difference is observed. This illustrates that the graph is able to capture the spatial derivatives.

\section{Conclusions}\label{sec:conclusions}
In this work, we have presented a novel data-driven framework for solving time-dependent nonlinear partial differential equations (PDE). The proposed approach is referred to as Graph Attention PDE or GrADE couples Feed-forward Neural Networks (FNN), Graph Attention (GAT), and Neural Ordinary Differential Equation (Neural ODE). The key idea is to use GAT to model the spatial domain an Neural ODE to model the temporal domain. FNNs are used for modeling the attention mechanism within the GAT network. GAT ensures that the problem at hand is computationally tractable as a node in the graph is only connected to its neighbors. Neural ODE, on the other hand, results in constant memory cost and allows trading of numerical precision for speed. While different numerical time-integration schemes can be used within the proposed framework, we have use forth order Runge Kutta method in this work. We also proposed depth refinement as an effective technique for training the proposed architecture in lesser time time with better accuracy.

We solve Burgers' equation to illustrate the performance of the proposed approach. Both 1D and 2D Burgers' equation has been solved. Results obtained have been benchmarked against those obtained using finite element solver. We observe that the proposed approach is able to provide accurate solution by using a larger time-step and snapshots of data at only two time-instants (referred as last integration time index). Case studies by varying last integration time index and amount of training data showcase the robustness of the proposed approach. We illustrated that the graph network layers are able to accurately capture the spatial derivatives of the state variables. We also showed that using depth refinement training strategy can help reduce training time for the network and increase its accuracy.


\textbf{Acknowledgements: } SC acknowledges the financial support received in form of seed grant from IIT Delhi.


\end{document}